%% file: arxiv.tex
\documentclass{article} 
\usepackage{iclr2027_conference,times}

\input{math_commands.tex}

\usepackage{hyperref}
\usepackage{url}

\usepackage{graphicx}
\usepackage{enumitem}
\usepackage{booktabs}
\usepackage{multirow}
\usepackage{algorithm}
\usepackage{algpseudocode}
\usepackage{amssymb}
\usepackage{tcolorbox}
\usepackage{amsthm}
\usepackage{mathtools}
\usepackage{wrapfig}   
\usepackage{makecell}  
\usepackage{colortbl}
\usepackage{empheq}
\usepackage{xcolor}
\usepackage{caption}
\usepackage{titletoc}

\theoremstyle{definition}
\newtheorem{property}{Property}

\definecolor{tableaccent}{HTML}{EDF3F8}

\makeatletter
\def\@oddhead{\vbox{\hbox{}\hfil\hrule height 0.4pt}}
\makeatother

\title{Naturalness-guided Manifold Flow Matching for Sign Language Production}

\author{Jiayi He $\quad$ Shengeng Tang$^{*}$ $\quad$ Sisi You $\quad$ Yanbin Hao $\quad$ Lechao Cheng $\quad$ Richang Hong \\
Hefei University of Technology \quad $^*$Corresponding author\\
}

\iclrfinalcopy 
\begin{document}

\maketitle

\begin{abstract}
Sign Language Production (SLP) aims to generate sign motions from text. Conditional Flow Matching methods have achieved strong performance in SLP by constructing conditional paths that transform a source distribution into a target distribution. However, existing methods construct these paths via linear interpolation, whereas the rotational geometry of human joints confines valid joint rotations to a manifold embedded in Euclidean space. Consequently, linear interpolation between two sign motions leaves this manifold and ignores the motion distribution on it. In this paper, we revisit SLP from the perspective of manifold transport and propose a Naturalness-guided Manifold Flow Matching framework, termed \textbf{SignNMFlow}, which constructs conditional paths directly on the motion manifold by jointly considering geometric efficiency and the motion distribution. Specifically, we exploit the intrinsic geometry of the manifold and introduce a motion naturalness measure to characterize the motion distribution. By minimizing the kinetic energy under this measure, we learn a naturalness-guided interpolation that couples a closed-form geodesic, which provides geometrically efficient transport, with a learnable deviation that incorporates the motion distribution, thereby significantly improving the fidelity of generated sign motions. Extensive qualitative and quantitative evaluations demonstrate the effectiveness of this work.

\end{abstract}

\section{Introduction}
Sign language is a visual-spatial language that conveys rich linguistic information through precise hand shapes and coordinated body movements. Existing research has primarily focused on Sign Language Recognition (SLR)~\citep{chen2022two, hu2023self, zuo2023natural, hu2023continuous} and Sign Language Translation (SLT)~\citep{chen2022two, zhou2023gloss, gong2024llms, wong2024sign2gpt, jang2025lost}, which aim to enable non-signers to understand sign language content. In contrast, Sign Language Production (SLP)~\citep{baltatzis2024neural, Zuo_2025_ICCV, tang2025sign, low2026signspark} has received relatively little attention. Unlike SLR and SLT, which decode existing sign language content, SLP must generate temporally coordinated motions that are semantically consistent with the text and visually natural.

\begin{figure*}[tbh]
    \centering
    \vspace{-3mm}
    \includegraphics[width=1.0\linewidth]{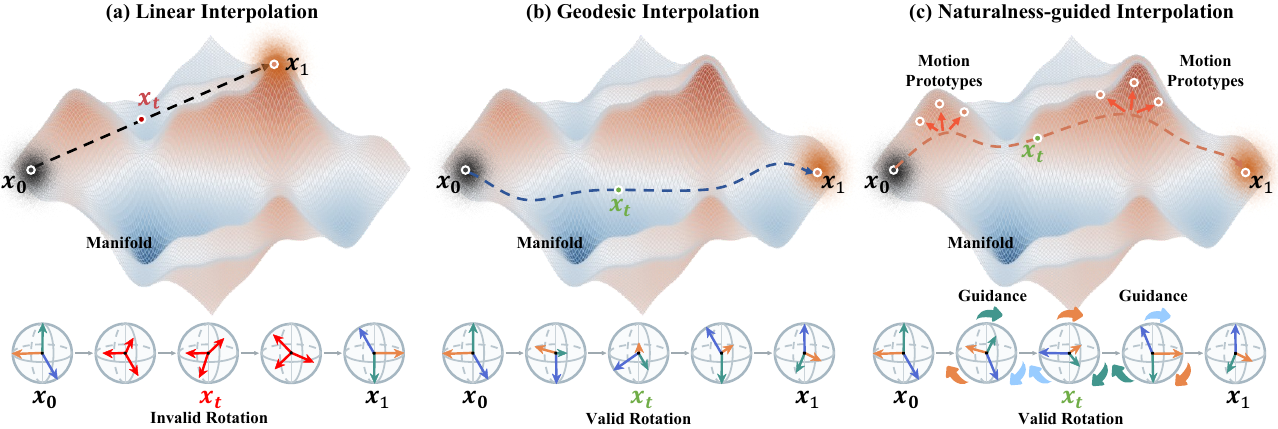}
    \vspace{-5mm}
    \caption{Here, $x_{0}$ and $x_{1}$ denote samples from the source and target distributions, and the color transition from \textcolor{blue}{blue} to \textcolor{orange}{orange} reflects the distribution of sign motions on the manifold. (a)~Linear interpolation leaves the manifold, yielding invalid rotations. (b)~Geodesic interpolation stays on the manifold but ignores the motion distribution. (c)~Our naturalness-guided interpolation stays on the motion manifold, maintaining geometric efficiency while incorporating the motion distribution.}
    \label{fig:motivation}
    \vspace{-3mm}
\end{figure*}

Existing SLP methods span three families: autoregressive models~\citep{saunders2020progressive, yin2024t2s, Zuo_2025_ICCV}, diffusion models~\citep{baltatzis2024neural, tang2025sign}, and flow-based generative models~\citep{khan2025signflow, low2026signspark}. Among these, Conditional Flow Matching (CFM)~\citep{lipmanflow} has demonstrated strong performance. CFM learns a velocity field that transports samples from a source distribution to the target distribution of sign motions, and anchors training on per-pair \emph{conditional paths}. Existing flow matching-based SLP methods~\citep{khan2025signflow, low2026signspark} construct these paths by linear interpolation, thereby ignoring the rotational geometry of human joints and the distribution of sign motions. As shown in Figure~\ref{fig:motivation}(a), linear interpolation between two sign motions introduces invalid intermediate rotations. This invalidity has a geometric origin. A joint rotation is a geometric object, and rotation representations merely embed rotations into Euclidean space~\citep{dunkel2024normalizing}, so the valid rotations of a sign motion reside on a structured manifold (Sec.~\ref{sec: issue}). As shown in Figure~\ref{fig:motivation}(b), a straight line between two sign motions in the ambient space therefore leaves the manifold, whereas its on-manifold counterpart, the geodesic, follows the intrinsic geometry. Moreover, linear interpolation depends only on the two endpoints and is independent of the distribution of sign motions. This geometric and distributional mismatch renders the learned velocity field inconsistent with the manifold.

A better construction of conditional paths should be established for flow matching-based SLP. Although \citet{sung2026context} demonstrated that geodesics outperform linear interpolation in optimizing motion transitions between glosses\footnote{In linguistics, glosses are minimal lexical items that match the meaning of signs.}, the improvement remains local to gloss transitions and leaves the geometric validity of the generative transport itself unexamined. We observe a geometric mismatch in existing flow matching-based SLP methods, where the conditional paths systematically deviate from the motion manifold while their construction ignores the motion distribution on it. Therefore, an ideal conditional path should stay on the manifold and ensure both geometric efficiency and consistency with the motion distribution, where geometric efficiency keeps the transport free of unnecessary detours and the motion
distribution enhances motion fidelity.

To this end, we propose a Naturalness-guided Manifold Flow Matching framework, termed SignNMFlow. As shown in Figure~\ref{fig:motivation}(c), SignNMFlow constructs conditional paths directly on the motion manifold by jointly considering geometric efficiency and the motion distribution. Specifically, we first extract prototypical motions from sign motions and introduce a motion naturalness measure to characterize the distribution of sign motions on the manifold. Building on this, we construct the conditional paths through the naturalness-guided interpolation learned by minimizing the kinetic energy under this naturalness measure. This interpolation couples a geometric term, the closed-form geodesic of the manifold that provides geometrically efficient
transport, with a naturalness-guided term, a learnable deviation from this geodesic that incorporates the motion distribution. In this way, SignNMFlow constructs geometrically efficient and motion-distribution-aware conditional paths, generating higher-fidelity sign motions. We summarize our \textbf{main contributions} as follows: 

\begin{enumerate}[label=(\arabic*), leftmargin=2.8em, itemsep=0em, topsep=0em]
    \item We revisit SLP from the perspective of manifold transport and identify the geometric mismatch issue (Sec.~\ref{sec: issue}) in existing flow matching-based SLP methods.
    \item We propose SignNMFlow (Sec.~\ref{sec: nmf}), a naturalness-guided manifold flow matching framework that constructs conditional paths directly on the motion manifold by jointly considering geometric efficiency and motion distribution.
    \item We introduce a motion naturalness measure (Sec.~\ref{sec: measure}) that characterizes the distribution of sign motions on the manifold and learn the naturalness-guided interpolation by minimizing the kinetic energy under this measure (Sec.~\ref{sec: path}).
    \item Extensive qualitative and quantitative evaluations on the Phoenix-2014T, CSL-Daily, and How2Sign datasets demonstrate the effectiveness of SignNMFlow, which achieves hand DTW-PA-JPE of $1.07$, $1.30$, and $2.07$, respectively.
\end{enumerate}

\section{Preliminaries}
In this section, we identify the geometric mismatch between linear interpolation and the motion manifold of sign motions (Sec.~\ref{sec: issue}), which motivates constructing conditional paths directly on the manifold. We then review flow matching (Sec.~\ref{sec: fm}), which forms the basis of SignNMFlow.

\subsection{Geometric Mismatch}
\label{sec: issue}
Constrained by the rotational geometry of human joints, valid sign motions do not occupy the full Euclidean space but reside on a manifold embedded within it. Formally, the rotation of each joint in a sign motion is represented by a rotation matrix $Q \in SO(3)$, where $SO(3)=\{Q \in \mathbb{R}^{3 \times 3} \mid Q^{\top}Q=I, \det(Q)=1\}$. Each frame of a sign motion therefore consists of a tuple of $J$ rotations, and the space of all such tuples forms the $J$-fold product manifold $SO(3)^J$, which is embedded in the Euclidean space $\mathbb{R}^{9J}$. However, existing flow matching-based SLP methods~\citep{khan2025signflow, low2026signspark} construct conditional paths by linearly interpolating rotation representations in this space. We demonstrate the inherent incompatibility between linear interpolation and rotational geometry by directly analyzing $SO(3)$. Given a source rotation $R_0\in SO(3)$ and a target rotation $R_1\in SO(3)$, the conditional path constructed by linear interpolation between them is defined as:
\begin{equation}
R_{t} = (1 - t)R_{0} + tR_{1}, \quad t \in [0, 1].
\label{eq: rot-path}
\end{equation}
Every $R\in SO(3)$ has orthonormal columns and hence satisfies $\Vert R \Vert_{F} = \sqrt{\mathrm{tr}(R^{\top}R)} = \sqrt{3}$. Expanding the squared Frobenius norm of the interpolated rotation via the Frobenius inner product $\langle \cdot, \cdot \rangle_{F}$ yields the closed form (see Appendix~\ref{app: sub-so3}):
\begin{equation}
\Vert R_{t} \Vert_{F}^{2} = 3 - t(1-t)\, \Vert R_{0} - R_{1} \Vert_{F}^{2}.
\label{eq: rot-norm}
\end{equation}
Whenever $R_{0}\neq R_{1}$, we have $\Vert R_{0}-R_{1}\Vert_{F}^{2} > 0$, so for any $t\in(0,1)$, $\Vert R_{t} \Vert_{F}^{2} < 3$.
The interpolated state $R_{t}$ therefore violates the orthogonality constraint that defines $SO(3)$, and linear interpolation inevitably produces intermediate states off the manifold. The transport adopted by existing flow matching-based SLP methods is thus misaligned with the intrinsic geometry of the motion manifold. We formalize this observation as Property~\ref{prop: geometric}.
\begin{property}
\label{prop: geometric}
\textit{Given two distinct rotations $R_0, R_1\in SO(3)$, the linear interpolation $R_{t}=(1-t)R_{0}+tR_{1}$, $t\in[0,1]$, produces intermediate states $R_{t} \notin SO(3)$ for all $t \in (0,1)$.}
\end{property}
In practice, joint rotations are commonly parameterized using the 6D rotation representation~\citep{zhou2019continuity}, unit quaternions, or Euler angles. Across these representations, linear interpolation fails to respect the underlying rotation geometry. This mismatch motivates constructing conditional paths directly on the motion manifold. For further details, please refer to Appendix~\ref{app: derivation}.

\subsection{Flow Matching}
\label{sec: fm}
Flow matching (FM)~\citep{albergobuilding} learns a velocity field that transports samples from a source distribution $p_{0}$ to a target distribution $p_{1}$ along a predefined probability path $p_{t}$, $t\in[0,1]$. Unlike score-based diffusion~\citep{Sohl_2015, Ho_Jain_Abbeel_Berkeley}, which learns the score function of a stochastic process, FM directly minimizes the discrepancy between a learnable vector field $v_{t}^{\theta}(x)$, given by a neural network with parameters $\theta$, and a target velocity field $u_{t}(x)$:
\begin{equation}
    \mathcal{L}_{\mathrm{FM}} = \mathbb{E}_{t\sim \mathcal{U}[0,1],\, x \sim p_{t}}\big[\Vert v_{t}^{\theta}(x) - u_{t}(x) \Vert^{2}\big].
\end{equation}
However, the target field $u_{t}(x)$ requires marginalizing over the full data distribution and is thus intractable in practice. Conditional Flow Matching (CFM)~\citep{lipmanflow} circumvents this issue by conditioning on individual source-target pairs. Given a source sample $x_{0}\sim p_{0}$ and a target sample $x_{1}\sim p_{1}$, one defines a conditional probability path $p_{t}(x|x_{0},x_{1})$ and a conditional velocity field $u_{t}(x| x_{0},x_{1})$. Common constructions include linear interpolation in Euclidean space~\citep{lipmanflow} and geodesic interpolation on manifolds~\citep{chen2024flow}. The CFM training objective can be stated as:
\begin{equation}
    \mathcal{L}_{\mathrm{CFM}} = \mathbb{E}_{t \sim \mathcal{U}[0,1],\, x_{0}\sim p_{0},\, x_{1}\sim p_{1}}\big[\Vert v_{t}^{\theta}(x_{t}) - u_{t}(x_{t}|x_{0}, x_{1}) \Vert^{2}\big],
    \label{eq: cfm}
\end{equation}
where $x_{t}$ is a sample from the conditional path at time $t$. This conditional objective is tractable and yields the same parameter gradients as $\mathcal{L}_{\mathrm{FM}}$, enabling efficient training through per-sample, path-based supervision.

\section{Naturalness-guided Manifold Flow Matching}
\label{sec: nmf}
We introduce \textsc{Naturalness-guided Manifold Flow Matching} (SignNMFlow), which constructs conditional paths directly on the motion manifold by jointly considering geometric efficiency and the motion distribution. SignNMFlow introduces a motion naturalness measure that characterizes the distribution of sign motions on the manifold and learns a naturalness-guided interpolation by minimizing the kinetic energy under this measure.

We organize this section as follows. In Sec.~\ref{sec: representation}, we present a manifold-based representation of sign motions, capturing the intrinsic rotational geometry of human joints. In Sec.~\ref{sec: measure}, we define the motion naturalness measure. In Sec.~\ref{sec: path}, we construct the naturalness-guided interpolation by minimizing the kinetic energy under this measure and train the sign motion generation network along the learned paths.

\subsection{Motion Representation}
\label{sec: representation}
Existing methods~\citep{khan2025signflow, low2026signspark} typically represent sign motions with the 6D rotation
representation~\citep{zhou2019continuity} and treat its coordinates as unconstrained Euclidean variables. Consequently, linear interpolation drives intermediate states off the manifold (Sec.~\ref{sec: issue}). To construct conditional paths directly on the motion manifold, we represent a sign motion $x$ as $J$ unit quaternions:
\begin{equation}
    x = \left( q_{1}, \dots, q_{J} \right) \in \mathcal{M} = (\mathbb{S}^{3})^{J},
    \label{eq: representation}
\end{equation}
where $J$ denotes the number of joints, and $\mathcal{M}$ is the $J$-fold product manifold of unit hyperspheres $\mathbb{S}^{3}$. Since $q$ and $-q$ represent the same rotation, we remove this sign ambiguity in Appendix~\ref{app: quaternion-sign}. A sign motion sequence with $T$ frames is represented as $\mathcal{X} = (x^{1}, \dots, x^{T})$. For notational clarity, we omit the frame index and develop the geometry on a single sign motion $x$ in the remainder of this paper. All definitions below extend to $\mathcal{X}$ by applying the same maps to each frame. For any point $x \in \mathcal{M}$, let $T_{x}\mathcal{M}$ denote the tangent space at $x$. The exponential map and the logarithmic map are defined as:
\begin{equation}
    \mathrm{Exp}_{x}: T_{x}\mathcal{M} \rightarrow \mathcal{M}, \qquad
    \mathrm{Log}_{x}: \mathcal{M} \rightarrow T_{x}\mathcal{M},
    \label{eq: explog}
\end{equation}
where $\mathrm{Exp}_{x}$ maps tangent vectors onto the manifold and $\mathrm{Log}_{x}$ is its local inverse.

\subsection{Motion Naturalness Measure}
\label{sec: measure}
A desirable conditional path on $\mathcal{M}$ should jointly consider geometric efficiency and the motion distribution. The former follows from the intrinsic geometry of $\mathcal{M}$, which Sec.~\ref{sec: path} exploits to construct the naturalness-guided interpolation. The latter requires an explicit characterization of the motion distribution on $\mathcal{M}$. To this end, we introduce a motion naturalness measure.

We first extract prototypical motions from the training sign motions via K-means clustering~\citep{mcqueen1967some}, obtaining $K$ prototypes $\{c_k\}_{k=1}^{K}$ with frequency weights $\{w_k\}_{k=1}^{K}$, where $w_k$ reflects the occurrence frequency of the rotations assigned to prototype $c_k$. The detailed procedure of the K-means-based prototype extraction is provided in Appendix~\ref{app: sub-prototype}. Given a motion $x\in\mathcal{M}$, we characterize the distribution of the prototypes around
$x$ through a kernel-weighted coefficient $\tilde{w}_k(x)$ and a local scatter matrix $S(x)$:
\begin{equation}
    \tilde{w}_k(x) = w_k \cdot \exp\left( -\frac{\|\delta_k\|^2}{2\sigma^2}
    \right), \quad
    S(x) = \sum_{k=1}^K \tilde{w}_k(x) \cdot \delta_k \delta_k^\top,
    \label{eq: scatter}
\end{equation}
where $\sigma$ is the kernel bandwidth and $\delta_k=\mathrm{Log}_{x}(c_k)\in T_x\mathcal{M}$ is the tangent vector from $x$ to prototype $c_k$. The kernel-weighted coefficient $\tilde{w}_k(x)$ quantifies the contribution of prototype $c_k$ to the motion distribution around $x$, combining its global frequency weight $w_k$ with a Gaussian proximity factor. The scatter matrix $S(x)$ complements this distributional information with the local dispersion of the prototypes around $x$, aggregating the outer products $\delta_k\delta_k^\top$ weighted by $\tilde{w}_k(x)$. We define the naturalness measure as:
\begin{equation}
    G(x) = m(x) \big( S(x) + \rho \mathbf{I} \big)^{-1},
    \label{eq: naturalness}
\end{equation}
where $m(x) = \sum_{k=1}^{K} \tilde{w}_k(x)$ denotes the aggregate kernel mass at $x$, and $\rho=1e-6$ is a regularization constant. We establish the validity of this measure in Appendix~\ref{app: sub-validity}. The point-wise measure $G(x)$ alone cannot evaluate an entire conditional path. We therefore lift it from points to paths, incorporating the motion distribution into path construction. For a path $\gamma:[0,1]\rightarrow\mathcal{M}$ with $\gamma_{0}=x_{0}$ and $\gamma_{1}=x_{1}$, we define its length $\mathcal{N}(\gamma)$ under the naturalness measure as:
\begin{equation}
    \mathcal{N}(\gamma) = \int_{0}^{1} \|\dot{\gamma}_t\|_{G(\gamma_t)} \,
    dt, \quad \|\dot{\gamma}_{t}\|_{G(\gamma_{t})} :=
    \sqrt{\langle \dot{\gamma}_{t}, G(\gamma_{t})^{-1} \dot{\gamma}_{t}
    \rangle},
    \label{eq: path-naturalness}
\end{equation}
where $\dot{\gamma}_{t}$ is the velocity of the path and $\langle \cdot, \cdot \rangle$ denotes the Euclidean inner product. 

\begin{algorithm}[t]
\caption{Training procedure of motion naturalness learning}
\label{alg: training}
\begin{algorithmic}[1]
\Require{motion manifold $\mathcal{M}$, source and target distributions
$p_{0}, p_{1}$, initialized correction network $\epsilon_{\theta}$,
naturalness measure $G(\cdot)$}

\While{Training}

    \State Sample $x_{0}\sim p_{0}$, $x_{1}\sim p_{1}$, and
    $t\sim\mathcal{U}[0,1]$

    \State $\xi=\mathrm{Log}_{x_0}(x_1)$,
    $\zeta(t)=\epsilon_{\theta}(t,x_0,x_1)$,
    $\omega(t)=t\,\xi+t(1-t)\,\zeta(t)$

    \State
    $x_{t, \theta}=\mathrm{Exp}_{x_0}(t\mathrm{Log}_{x_0}(x_1)+t(1-t)\epsilon_{\theta}(t,x_0,x_1))$
    \Comment{Eq.~\eqref{eq: learnable-path}}

    \State
    $\dot{x}_{t,\theta}
    =
    \left(D\mathrm{Exp}_{x_0}\right)_{\omega(t)}
    \left[
    \xi+(1-2t)\,\zeta(t)
    +
    t(1-t)\,\frac{\partial}{\partial t}\epsilon_{\theta}(t,x_0,x_1)
    \right]$

    \State
    $\ell(\theta)
    \leftarrow
    (\dot{x}_{t,\theta})^{\top}
    G(x_{t,\theta})^{-1}
    \dot{x}_{t,\theta}$
    \Comment{Estimate of $\mathcal{L}_{G}(\theta)$ in Eq.~\eqref{eq: ln}}

    \State Update $\theta$ using gradient $\nabla_{\theta}\ell(\theta)$

\EndWhile

\State \textbf{return} naturalness-guided transport paths parameterized by
$\epsilon_{\theta}$
\end{algorithmic}
\end{algorithm}

\subsection{Motion Naturalness Learning}
\label{sec: path}
With the motion representation (Sec.~\ref{sec: representation}) and the motion naturalness measure (Sec.~\ref{sec: measure}) established on $\mathcal{M}$, we construct conditional paths that secure geometric efficiency through the intrinsic geometry of $\mathcal{M}$, and incorporate the motion distribution by minimizing the kinetic energy under this measure. We then train the sign
motion generation network along the learned paths.

\textbf{Learning the naturalness-guided interpolation.}
Given a source sample $x_{0} \sim p_{0}$ and a target sign motion sample $x_{1} \sim p_{1}$, an ideal conditional path $\gamma:[0,1]\rightarrow \mathcal{M}$ should satisfy the boundary conditions $\gamma_{0}=x_{0}$ and $\gamma_{1}=x_{1}$. In differential geometry, the geodesic follows the intrinsic geometry of $\mathcal{M}$ and provides the shortest connection between endpoints. However, the geodesic depends solely on the geometry of $\mathcal{M}$ and ignores the motion distribution. To incorporate the motion distribution, we define the naturalness-guided interpolation as the conditional path that minimizes the kinetic energy under the naturalness measure:
\begin{equation}
    \gamma^{\star} = \underset{\gamma:\,\gamma_0=x_0,\;\gamma_1=x_1}
    {\arg\min} \, \varphi(\gamma), \quad
    \varphi(\gamma) := \mathbb{E}_{t \sim \mathcal{U}[0,1]}
    \left[ \dot{\gamma}_{t}^{\top} G(\gamma_{t})^{-1}\dot{\gamma}_{t}\right],
    \label{eq: psi}
\end{equation}
where $\varphi$ is the kinetic energy under the naturalness measure, i.e., the energy form of $\mathcal{N}(\gamma)$ in Eq.~\ref{eq: path-naturalness}. If the naturalness measure were uniform, the minimizer of $\varphi$ would be exactly the geodesic, so any deviation of $\gamma^{\star}_{t}$ from the geodesic is driven by the non-uniform motion distribution. Since the exact minimizer over all admissible paths is intractable, we parameterize the interpolation around the geodesic with a learnable correction network $\epsilon_{\theta}$, whose output is orthogonally projected onto the tangent space $T_{x_0}\mathcal{M}$ (the projection operator and the exactness guarantee are provided in Appendix~\ref{app: sub-proof}):
\begin{equation}
    x_{t, \theta}=\mathrm{Exp}_{x_0}(\underbrace{t\cdot
    \mathrm{Log}_{x_{0}}(x_1)}_{\mathrm{Geometric}} +
    \underbrace{t(1-t)\cdot\epsilon_{\theta}(t,x_0,x_1)}_{\mathrm{Naturalness}}),
    \quad t\in[0,1],
    \label{eq: learnable-path}
\end{equation}
where the geometric term $t\cdot\mathrm{Log}_{x_0}(x_1)$ is the closed-form geodesic that provides geometrically efficient transport, and the naturalness term $t(1-t)\cdot\epsilon_{\theta}(\cdot)$ is a learnable deviation from it, trained by minimizing $\varphi$. The factor $t(1-t)$ vanishes at both endpoints, so the boundary conditions hold for any $\epsilon_{\theta}$. Minimizing $\varphi(\gamma)$ over the parameterized path $x_{t, \theta}$ yields our training objective (the training procedure is summarized in Algorithm~\ref{alg: training}):
\begin{tcolorbox}[
    colback=gray!10,
    colframe=gray!10,
    boxrule=0pt,
    left=5pt,
    right=5pt,
    top=5pt,
    bottom=5pt
]
\begin{equation}
    \mathcal{L}_{G}(\theta):=
    \mathbb{E}_{t\sim\mathcal{U}[0,1],\; x_{0}\sim p_{0},\; x_{1}\sim p_{1}}
    [
    \dot{x}_{t,\theta}^{\top}
    G(x_{t,\theta})^{-1}
    \dot{x}_{t,\theta}
    ],
    \label{eq: ln}
\end{equation}
\end{tcolorbox}
where $\dot{x}_{t,\theta}$ denotes the velocity of $x_{t,\theta}$ at time $t$. In practice, we compute $\dot{x}_{t,\theta}$ by automatic differentiation. The chain-rule expression in Algorithm~\ref{alg: training} is its analytic equivalent.

\begin{table}[htbp]
\caption{Comparisons with state-of-the-art sign language production (text-to-sign) methods. Best mean values are in \textbf{bold} and second-best are \underline{underlined}.}
\vspace{-3mm}
\centering
\label{tab:metrics}
\resizebox{\textwidth}{!}{
\begin{tabular}{l ccccc  ccccc}
\toprule
  \multirow{3}{*}{Methods} 
& \multicolumn{5}{c}{Phoenix-2014T} 
& \multicolumn{5}{c}{CSL-Daily} \\
& \multicolumn{2}{c}{DTW-JPE} & \multicolumn{2}{c}{DTW-PA-JPE} & B-T 
& \multicolumn{2}{c}{DTW-JPE} & \multicolumn{2}{c}{DTW-PA-JPE} & B-T \\
\cmidrule(lr){2-3} \cmidrule(lr){4-5} \cmidrule(lr){6-6} \cmidrule(lr){7-8} \cmidrule(l){9-10} \cmidrule(lr){11-11} 
& Body & Hand & Body & Hand & BLEU-4 & Body & Hand & Body & Hand & BLEU-4 \\
\midrule
Pr.~Tr.~\citep{saunders2020progressive}
& 15.01 & 31.77 & 13.67 & 11.95 & 4.94
& 16.30 & 32.63 & 15.98 & 12.91 & 3.07 \\

T2Mesh~\citep{stoll2022there}
& 14.04 & 31.64 & 13.48 & 12.06 & 5.81
& 13.76 & 30.37 & 13.47 & 12.10 & 5.11 \\

T2S-GPT~\citep{yin2024t2s}
& 11.65 & 19.09 & 10.38 & 6.47 & 9.06
& 12.32 & 15.43 & 11.94 & 5.93 & 8.94 \\

S-MGPT~\citep{jiang2023motiongpt}
& 10.42 & 9.08  & 9.45  & 3.41 & 9.68
& 11.58 & 11.31 & 10.81 & 3.78 & 8.82 \\

SignFlow~\citep{khan2025signflow}
& 5.40 & 6.21 & 4.56 & 1.20 & --
& 7.56 & 8.82 & 6.43 & 1.65 & -- \\

SOKE~\citep{Zuo_2025_ICCV} 
& 6.04 & 7.72 & 4.77 & 1.38 & 11.87
& 7.38 & 9.68 & 6.24 & 1.71 & 11.30 \\

SignSparK~\citep{low2026signspark}
& \underline{4.40} & 7.10  & 5.24 & 1.52 & --
& 6.27 & 10.63 & 7.26 & 2.00 & -- \\

\midrule
SignNMFlow
& \textbf{4.37} & \textbf{5.57} & \textbf{3.85} & \textbf{1.07} & \textbf{16.15}
& \textbf{5.73} & \textbf{7.43} & \textbf{5.10} & \textbf{1.30} & \underline{15.14}  \\

SignNMFlow-M
& 4.42 & \underline{5.75} & \underline{3.91} & \underline{1.08} & \underline{14.45}
& \underline{5.94} & \underline{7.73} & \underline{5.31} & \underline{1.33} & \textbf{15.23}\\
\bottomrule
\end{tabular}}
\vspace{-3mm}
\end{table}

\textbf{Training generation along the learned paths.} 
With the naturalness-guided interpolation $x_{t,\theta}$, we freeze the correction network $\epsilon_{\theta}$ and train the sign motion generation network $\phi$ along this path. Rather than directly regressing the conditional velocity field as in standard CFM Eq.~\ref{eq: cfm}, we adopt a \emph{target-prediction} parameterization. The generation network $\phi$ takes the intermediate state
$x_{t}$ along the learned path, the paired expression state $e_{t}$, the time step $t$, and the text condition $c$ as input, and predicts the raw target motion $\tilde{x}_{1}$ together with the target expression $\hat{e}_{1}$. The motion prediction is normalized per joint onto $\mathcal{M}$ through the radial projection $\hat{x}_{1}^{j} = \tilde{x}_{1}^{j}/\Vert\tilde{x}_{1}^{j}\Vert$, $j = 1, \dots, J$. Formally, the training objective for the motion channel is
\begin{tcolorbox}[
    colback=gray!10,
    colframe=gray!10,
    boxrule=0pt,
    left=5pt,
    right=5pt,
    top=5pt,
    bottom=5pt
]
\begin{equation}
    \mathcal{L}_{\mathrm{NMF}}(\phi) =
    \mathbb{E}_{t \sim \mathcal{U}[0,1],\; x_0 \sim p_0,\; x_1 \sim p_1}
    \big[ \Vert x_1 - \hat{x}_{1} \Vert^{2} \big].
    \label{eq: lnmf}
\end{equation}
\end{tcolorbox}
The facial expression channel is trained in Euclidean space under the same objective. At sampling time, each solver step re-predicts the target $\hat{x}_{1}$ from the current state, re-evaluates the correction with this prediction, and advances the state to the next path point of Eq.~\ref{eq: learnable-path}. The complete procedures are given in Appendix~\ref{app: sub-generation}. The transport underlying generation is thus both geometrically efficient and aware of the motion distribution, yielding higher-fidelity sign motions.

\textbf{Geometric efficiency and naturalness on equal footing.} Geometric efficiency and naturalness differ in origin. The former arises from the intrinsic geometry of $\mathcal{M}$ and enters the path through the fixed geodesic, whereas the latter arises from the motion distribution and enters the path through the learnable deviation optimized by the kinetic energy objective. Naturalness itself is defined in the same geometric terms, with both the scatter matrix $S(x)$ and the kinetic energy expressed in tangent spaces via the logarithmic map. In the learned path, the geodesic remains the fixed reference while the naturalness term absorbs the deviation induced by the motion distribution, so neither consideration is obtained at the expense of the other.

\section{Experiments}
\subsection{Experimental Settings}
\label{subsec:settings}

\textbf{Datasets.} Following existing methods~\citep{Zuo_2025_ICCV, low2026signspark}, we evaluate our method on three sign language datasets: Phoenix-2014T~\citep{camgoz2018neural}, CSL-Daily~\citep{zhou2021improving}, and How2Sign~\citep{duarte2021how2sign}, covering German, Chinese, and American sign language, respectively. For sign motion extraction, we adopt SMPL-X~\citep{pavlakos2019expressive} parameters, which are sourced from NSA~\citep{baltatzis2024neural} for How2Sign and from SOKE~\citep{Zuo_2025_ICCV} for Phoenix-2014T and CSL-Daily. SMPL-X represents each sign motion with joint rotations and facial expression parameters. Our manifold construction applies to the joint rotations, while the expression parameters are generated alongside them in Euclidean space under the same target-prediction objective as Eq.~\ref{eq: lnmf} (Appendix~\ref{app: sub-generation}). SignNMFlow is trained on a single dataset, and its multilingual version, SignNMFlow-M, is trained on the combination of all three, following the training setup of SOKE~\citep{Zuo_2025_ICCV}. 

\begin{figure*}[tbh]
    \centering
    \vspace{-4mm}
    \includegraphics[width=1.0\linewidth]{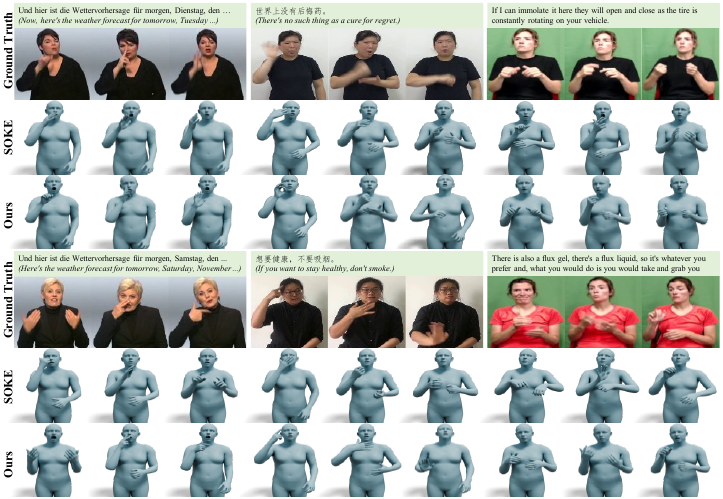}
    \vspace{-5mm}
    \caption{Qualitative comparisons of generated sign motions between our method with the SOKE, on the test sets of Phoenix-2014T(left), CSL-Daily(middle), and How2Sign(right).}
    \label{fig:visualization}
    \vspace{-5mm}
\end{figure*}

\textbf{Evaluation Metrics.} To evaluate our performance, we follow SOKE~\citep{Zuo_2025_ICCV}, using Dynamic Time Warping on the joint positions (DTW-JPE)  and its Procrustes-aligned version (DTW-PA-JPE) to assess the accuracy of the sign motions. Furthermore, we used BLEU-4 scores calculated through Back-Translation to assess the semantic intelligibility of the generated sign motions. To ensure fairness, our Back-Translation model is the same as SOKE. 

\textbf{Implementation Details.} For the correction network $\epsilon_{\theta}$, we empirically use a 3 layers MLP with a hidden dimension of 512. We train the correction network $\epsilon_{\theta}$ with a batch size of 8 for 100 epochs, using the AdamW optimizer~\citep{loshchilov2017decoupled} and a cosine learning rate scheduler starting at 2e-4. For the text encoder, we use the pre-trained mBART-large-cc25~\citep{liu2020multilingual}. For the sign motion generation network $\phi$, we use a transformer decoder network with the 8 decoder layers containing a separate cross-attention head for text. The timestep $t$ is passed to the decoder layers through Stylization block~\citep{zhang2024motiondiffuse} after every self-attention, cross-attention, feed-forward layer. We train the denoising network with a batch size of 64 per GPU for 300 epochs, employing the same optimizer settings as the neural network $\epsilon_{\theta}$. All models are trained on 4 NVIDIA A40 GPUs. Further details are provided in Appendix~\ref{app: exp det}.

\subsection{Comparison with State-of-the-Art Methods}

\begin{wraptable}{r}{0.6\textwidth}
\vspace{0mm}
\caption{Comparisons with state-of-the-art sign language production methods (text-to-sign) on How2Sign Dataset.}
\vspace{-3mm}
\centering
\label{tab:metrics1}
\resizebox{0.6\textwidth}{!}{
\begin{tabular}{l ccccc}
\toprule
\multirow{2}{*}{Methods} 
& \multicolumn{2}{c}{DTW-JPE} & \multicolumn{2}{c}{DTW-PA-JPE} & B-T   \\
\cmidrule(lr){2-3} \cmidrule(lr){4-5} \cmidrule(lr){6-6}
& Body & Hand & Body & Hand & BLEU-4 \\
\midrule
Pr.Tr. 
& 14.74 & 30.17 & 14.15 & 11.57 & 2.75 \\
T2Mesh 
& 15.50 & 32.97 & 13.99 & 13.47 & 7.51 \\
T2S-GPT
& 12.65 & 18.44 & 11.48 & 6.39  & 11.20 \\
S-MGPT 
& 12.41 & 13.74 & 11.23 & 4.39  & 11.45 \\
SignFlow 
& 7.98 & 10.52 & 6.92 & 2.27  & -- \\
SOKE
& 7.75  & 10.08 & 6.82  & 2.35  & 14.48 \\
SignSparK 
& \textbf{6.30} & 11.43 & 7.26  & 2.72 & -- \\
\midrule
SignNMFlow
& 6.76 & \textbf{9.72} & \underline{6.01} & \underline{2.07} & \underline{16.88}  \\
SignNMFlow-M
& \underline{6.74} & \underline{9.77} & \textbf{5.99} & \textbf{2.06} & \textbf{17.49} \\
\bottomrule
\end{tabular}}
\vspace{-3mm}
\end{wraptable}

\paragraph{Quantitative Comparison.} In Table~\ref{tab:metrics} and Table~\ref{tab:metrics1}, we compare SignNMFlow with state-of-the-art gloss-free SLP methods, focusing on three representative methods: SignFlow~\citep{khan2025signflow}, SOKE~\citep{Zuo_2025_ICCV}, and SignSparK~\citep{low2026signspark}. SOKE leverages a pre-trained language model with retrieval enhancements to improve SLP semantics; however, SignNMFlow achieves superior semantic fidelity despite these augmentations. SignFlow and SignSparK are both conditional flow matching frameworks that construct conditional paths by linear interpolation in Euclidean space, with SignFlow using optimal transport and SignSparK adopting sparse keyframes-based learning. Notably, SignNMFlow consistently outperforms both, achieving significantly lower hand DTW-PA-JPE errors of $1.07$, $1.30$, and $2.07$ on Phoenix-2014T, CSL-Daily, and How2Sign, respectively. Meanwhile, the multilingual version of SignNMFlow, SignNMFlow-M, has also demonstrated strong competitiveness. These results demonstrate the advantage of constructing conditional paths directly on the motion manifold.

\paragraph{Qualitative Comparison.} In Figure~\ref{fig:visualization}, we present qualitative comparisons between SignNMFlow and SOKE~\citep{Zuo_2025_ICCV}. Our method can continuously generate more realistic and expressive sign motions and significantly improve the fidelity of gestures. The advantage of SignNMFlow lies in its ability to construct conditional paths directly on the motion manifold, thereby more effectively preserving the intrinsic geometric structure of sign motions. Unlike SOKE, which may produce intermediate motion deviations, SignNMFlow constructs conditional paths through the naturalness-guided interpolation, generating sign motions that are significantly closer to the ground truth and demonstrating higher pose fidelity and spatial consistency. This highlights the effectiveness of SignNMFlow in capturing complex sign motion dynamics and generating high-fidelity sign motions.

\begin{figure}[t]
\centering
\begin{minipage}[t]{0.48\textwidth}
\centering
\vspace{-3pt}
\includegraphics[width=\linewidth]{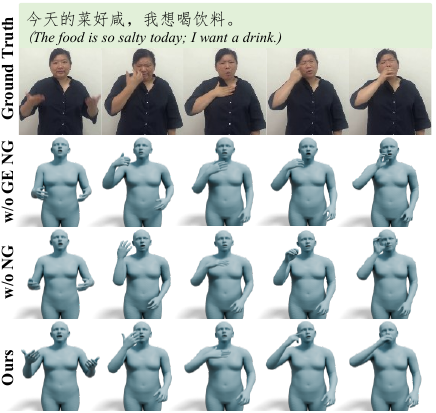}
\vspace{-3mm}
\caption{Qualitative ablation for GE and NG.}
\vspace{-3mm}
\label{fig:visualization ablation}
\end{minipage}
\hfill
\begin{minipage}[t]{0.48\textwidth}
\centering
\vspace{-3pt}
\captionof{table}{Ablation results of Motion Representation (MR), Geometric Efficiency (GE) and Naturalness-Guided (NG).}
\vspace{-3mm}
\label{tab:ablation}
\resizebox{\linewidth}{!}{
\begin{tabular}{ccccccccc}
\toprule
\multicolumn{3}{c}{Modules} &
\multicolumn{3}{c}{DTW-JPE$\downarrow$} &
\multicolumn{3}{c}{DTW-PA-JPE$\downarrow$}\\
\cmidrule(lr){4-6} \cmidrule(lr){7-9}
MR &GE & NG & All & Body & Hand & All & Body & Hand\\
\midrule
$\times$ &$\times$ &$\times$  
& 20.08 & 6.70 & 9.25 & 10.78 & 6.07 & 1.68 \\
$\checkmark$ &$\times$ &$\times$  
& 21.25 & 6.84 & 9.73 & 11.26 & 6.24 & 1.80 \\
$\checkmark$ &$\checkmark$ &$\times$
& 18.31 & 6.14 & 8.05 & 9.58 & 5.51 & 1.41 \\
$\checkmark$ &$\checkmark$ &$\checkmark$
& \textbf{16.86} & \textbf{5.73} & \textbf{7.43} 
& \textbf{8.92} & \textbf{5.10} & \textbf{1.30} \\
\bottomrule
\end{tabular}}
\vspace{2mm}

\captionof{table}{Ablation results of Measure.}
\label{tab:ablation measure}
\resizebox{\linewidth}{!}{
\begin{tabular}{cc ccc ccc}
\toprule
\multicolumn{2}{c}{\multirow{2}{*}{Parameters}} &
\multicolumn{3}{c}{DTW-JPE$\downarrow$} &
\multicolumn{3}{c}{DTW-PA-JPE$\downarrow$}\\
\cmidrule(lr){3-5} \cmidrule(lr){6-8}
\multicolumn{2}{c}{} 
& All & Body & Hand & All & Body & Hand\\
\midrule
\multicolumn{1}{c|}{\multirow{3}{*}{$K$}}  
& 250 
& 17.17 & 5.83 & 7.49
& \textbf{8.75}  & 5.19 & 1.32 \\

\multicolumn{1}{c|}{}
& 500 
& \textbf{16.86} & \textbf{5.73} & \textbf{7.43} 
& 8.92 & \textbf{5.10} & 1.30 \\

\multicolumn{1}{c|}{}
& 750 
& 17.32 & 5.87 & 7.53
& 8.80 & 5.20 & \textbf{1.29} \\

\midrule

\multicolumn{1}{c|}{\multirow{3}{*}{$\sigma$}}  
& 0.25 
& 19.41 & 6.47 & 8.71 
& 10.03 & 5.84 & 1.57 \\

\multicolumn{1}{c|}{}
& 0.50 
& \textbf{16.86} & \textbf{5.73} & \textbf{7.43} 
& 8.92 & \textbf{5.10} & \textbf{1.30} \\

\multicolumn{1}{c|}{}
& 0.75 
& 17.58 & 5.95 & 7.57 
& \textbf{8.80}  & 5.28 & 1.32 \\

\bottomrule
\end{tabular}}
\vspace{-2mm}
\end{minipage}
\vspace{-2mm}
\end{figure}

\subsection{Ablation Study}
In this subsection, we present ablation results to verify the effectiveness of SignNMFlow. All results are evaluated on CSL-Daily, while Phoenix-2014T and How2Sign are not used for ablations.

\textbf{Geometric Efficiency and Naturalness-Guided Analysis.} In Table~\ref{tab:ablation}, we present the ablation results for Geometric Efficiency (GE) and Naturalness-Guided (NG). All variants are flow-matching SLP models sharing the same \emph{target-prediction} parameterization. The baseline (first row) uses the common 6D rotation representation and constructs its conditional paths by linear interpolation in Euclidean space. Switching to the quaternion-based Motion Representation (MR) alone slightly raises DTW-JPE All error from $20.08$ to $21.25$. Linear interpolation treats quaternion components as flat coordinates rather than points on the rotation manifold, so the representation change alone brings no gain under this path geometry. GE removes this mismatch by constructing conditional paths as geodesics on the manifold instead of by linear interpolation. Paired with MR, GE lowers DTW-JPE All error from $21.25$ to $18.31$ and DTW-PA-JPE All error from $11.26$ to $9.58$, both below the 6D baseline ($20.08$ and $10.78$). The quaternion representation becomes superior only once the path geometry agrees with the manifold on which rotations lie. In addition, NG considers the motion distribution on the manifold. It brings further gains, reducing the two metrics to $16.86$ and $8.92$, the best results across all variants and $16.04\%$ and $17.25\%$ below the baseline, with consistent improvements on body and hand joints. Furthermore, we perform a qualitative comparison in Figure~\ref{fig:visualization ablation} by removing NG (w/o NG), and further removing both NG and GE (w/o GE NG). The figure demonstrates that NG and GE effectively improve the quality and fidelity of generated motions.

\textbf{Measure Configuration.} In Table~\ref{tab:ablation measure}, we present the ablation results regarding the parameter configurations for the motion naturalness measure, where $K$ is the number of motion prototypes and $\sigma$ is the kernel bandwidth in Eq.~\ref{eq: scatter}. For $K$, performance peaks around $500$ and mildly degrades at $750$. Too few prototypes underfit the motion distribution, while too many over-fragment it into small clusters whose frequency weights become unreliable. For $\sigma$, the effect is more pronounced. With $\sigma=0.25$, the kernel weights $\tilde{w}_{k}(x)$ concentrate on the single nearest prototype, so $G(x)$ tracks local prototype fluctuations rather than the overall distribution, and the DTW-JPE All error rises from $16.86$ to $19.41$. With $\sigma=0.75$, the kernel over-smooths the prototypes, so the weights become nearly uniform and $G(x)$ no longer reflects the non-uniform motion distribution. The interpolation accordingly loses part of its distributional guidance, and the DTW-JPE All error rises from $16.86$ to $17.58$, a milder degradation than at $\sigma=0.25$. We therefore use $K=500$ and $\sigma=0.50$.

\begin{wraptable}{r}{0.65\textwidth} 
\centering 
\vspace{-3mm}
\caption{Ablation results of training objectives.} 
\vspace{-3mm} 
\label{tab:ablation objective} 
\resizebox{0.65\textwidth}{!}{ \begin{tabular}{cc ccc ccc} 
\toprule 
\multirow{2}{*}{Method}  
& \multirow{2}{*}{Prediction}
& \multicolumn{3}{c}{DTW-JPE$\downarrow$} 
& \multicolumn{3}{c}{DTW-PA-JPE$\downarrow$}\\ 
\cmidrule(lr){3-5} \cmidrule(lr){6-8} 
& & All & Body & Hand & All & Body & Hand\\ 
\midrule 
\multirow{2}{*}{CFM} 
& Velocity & 21.62 & 7.43 & 9.49 & 11.33 & 6.64 & 1.79 \\ 
& Target   & \textbf{20.08} & \textbf{6.70} & \textbf{9.25} & \textbf{10.78} & \textbf{6.07} & \textbf{1.68} \\ 
\midrule 
\multirow{2}{*}{Ours} 
& Velocity & 18.76 & 6.44 & 8.13 & 9.52 & 5.69 & 1.44 \\ 
& Target & \textbf{16.86} & \textbf{5.73} & \textbf{7.43} & \textbf{8.92} & \textbf{5.10} & \textbf{1.30} \\
\bottomrule 
\end{tabular}} 
\vspace{-3mm} 
\end{wraptable}

\textbf{Training Objective.} In Table~\ref{tab:ablation objective}, we compare \emph{velocity-prediction} and \emph{target-prediction} parameterizations of the training objective under both the CFM baseline and our method. For the CFM model, target prediction reduces DTW-JPE All error from $21.62$ to $20.08$ and DTW-PA-JPE All error from $11.33$ to $10.78$. By contrast, our method reduces these two metrics from $18.76$ to $16.86$ and from $9.52$ to $8.92$, respectively, with consistent performance improvements on both body and hand joints. The advantage stems from the regression target itself. Velocity prediction regresses the difference between the sampled noise and the data, and the scale of this target varies across samples. Target prediction instead regresses the ground-truth motion directly, and its supervision signal remains well scaled across timesteps.

\subsection{User Study}
\label{sub-sec: user study}

\begin{wrapfigure}{r}{0.65\textwidth} 
\centering 
\vspace{-3mm}
\includegraphics[width=\linewidth]{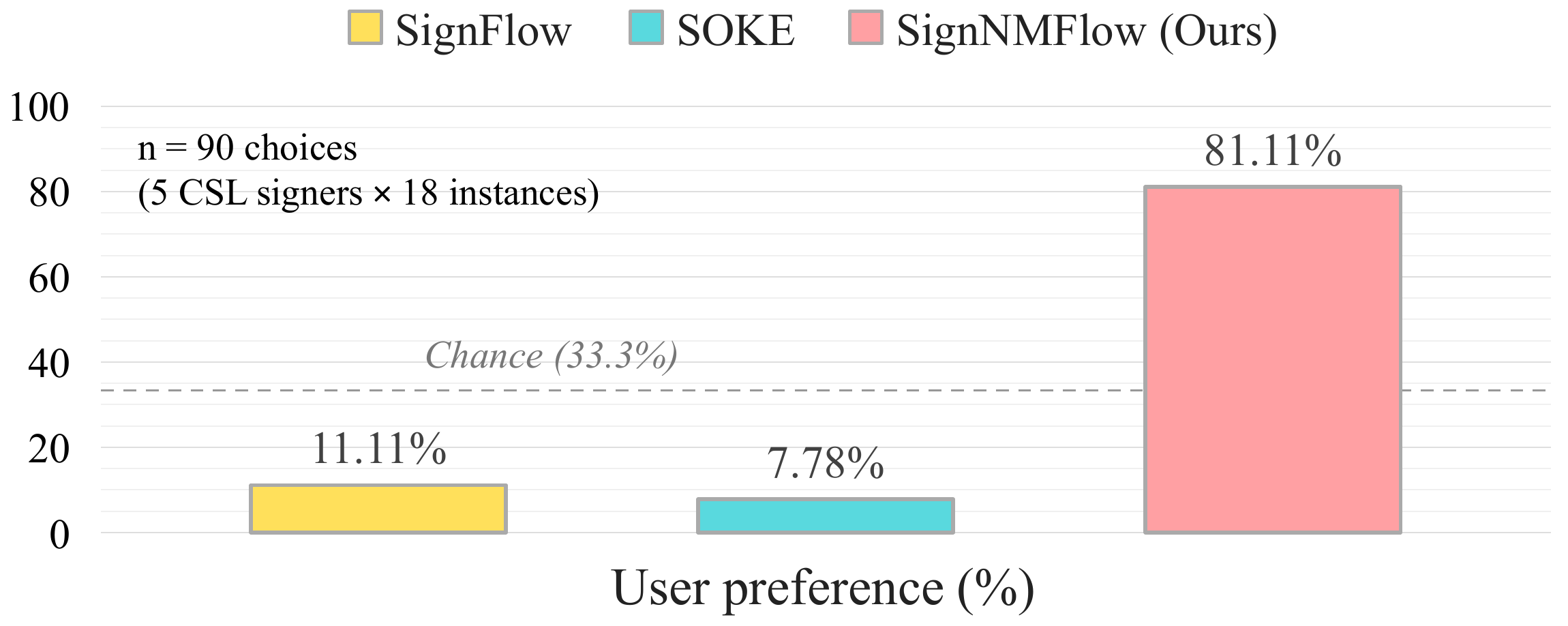} 
\vspace{-5mm} 
\caption{User study on the CSL-Daily: Ours vs. SOTA methods.} 
\label{fig:user study} 
\vspace{-3mm}
\end{wrapfigure} 
In addition to objective metrics, we invited 5 professional CSL signers to participate in a user study. These signers made a forced choice based on the semantic alignment between the generated sign motions and text annotations, together with the fidelity of the motion. Higher user preference scores are associated with better semantic alignment and higher-fidelity motion. Specifically, we provided 18 generated sign motions from the SOTA baselines (SignFlow and SOKE) and SignNMFlow. The order of the sign motions was randomly shuffled to prevent potential bias. As shown in Figure~\ref{fig:user study}, SignNMFlow was preferred in $81.11\%$ of the trials, significantly outperforming SOKE ($7.78\%$) and SignFlow ($11.11\%$). These results are consistent with qualitative and quantitative evaluations, demonstrating the improvements of SignNMFlow in semantic alignment and motion fidelity.

\section{Conclusions}
In this paper, we revisit SLP from the perspective of manifold transport and propose SignNMFlow, a naturalness-guided manifold flow matching framework. Unlike existing flow matching-based SLP methods that construct conditional paths by linear interpolation, SignNMFlow constructs conditional paths directly on the motion manifold, jointly considering geometric efficiency and the motion distribution. Specifically, we represent sign motions on the product of unit-quaternion hyperspheres and construct the conditional paths through the naturalness-guided interpolation, which couples a closed-form geodesic with a learnable deviation. To incorporate the motion distribution, we introduce a motion naturalness measure built from prototypical motions and learn the interpolation by minimizing the kinetic energy under this measure. Additionally, we adopt a target-prediction parameterization that provides a well-scaled supervision signal without extra losses. Extensive evaluations on Phoenix-2014T, CSL-Daily, and How2Sign validate the effectiveness of SignNMFlow.

\subsection*{AI use statement}
In this work, we have not used generative AI tools for any task that requires disclosure, including generating synthetic data, developing theoretical models or conceptual frameworks, formulating or proving mathematical claims, proposing hypotheses, designing or providing feedback on research methodology or experiments, implementing methods, or interpreting results. We used generative AI tools only to edit the manuscript and improve the readability of the English text. We have reviewed all AI-assisted text and verified that the technical content, claims, and reported numbers originate from our own experiments and derivations. We take responsibility for the final content of this work, including text, claims, or artifacts produced with the aid of generative AI.

\subsection*{Ethics statement}
This work studies sign language production, whose long-term goal is to improve accessibility for Deaf and hard-of-hearing communities; the generated sign motions are intended to support, not replace, human interpreters. All experiments rely on three publicly released research benchmarks, Phoenix-2014T, CSL-Daily, and How2Sign, which are used for their intended research purpose, and no new signer data are collected for training. The user study involved five CSL signers who participated voluntarily with informed consent; the study records only aggregate selection counts over anonymized generated motions, and no personally identifiable information is collected or reported.

\subsection*{Reproducibility statement}
All experiments are conducted on three public benchmarks, Phoenix-2014T, CSL-Daily, and How2Sign; the datasets, evaluation metrics, and evaluation protocol are described in Sec.~\ref{subsec:settings}. The training and sampling procedures of SignNMFlow are summarized in Algorithm~\ref{alg: training} and Algorithm~\ref{alg: generation}. Complete architecture and training hyperparameters, including the configuration of the motion naturalness measure, are listed in Appendix~\ref{app: exp det}, and the full derivation of the geometric mismatch (Property~\ref{prop: geometric}) is provided in Appendix~\ref{app: derivation}. The comparison with prior work follows the protocol of SOKE~\citep{Zuo_2025_ICCV}. 

\bibliography{iclr2027_conference}
\bibliographystyle{iclr2027_conference}

\newpage
\appendix
{\centering
\LARGE\textbf{Appendix}
\par}
\vspace{5mm}

\startcontents[appendix]

\titlecontents{section}[2em]      
    {\vspace{1.5mm}\bfseries}
    {\thecontentslabel\hspace{0.8em}}
    {}
    {\hfill\bfseries\thecontentspage}
\titlecontents{subsection}[3.8em] 
    {\vspace{0.6mm}}
    {\thecontentslabel\hspace{0.5em}}
    {}
    {\titlerule*[0.6em]{.}\thecontentspage}

\noindent{\large\textbf{Table of Contents}}
\par
\vspace{-5mm}
\noindent\rule{\textwidth}{0.4pt}
\vspace{-5mm}
\setcounter{tocdepth}{2}
\printcontents[appendix]{}{1}{}
\noindent\rule{\textwidth}{0.4pt}
\newpage

\section{Related Work}
\subsection{Sign Language Production} 
Early SLP methods~\citep{glauert2006vanessa, karpouzis2007educational} primarily relied on rule-based systems and manually designed animation libraries, which achieved text-to-sign conversion by establishing mappings between textual semantics and predefined sign motions. However, due to their dependence on manually crafted rules and limited motion templates, these methods struggled to model the complex semantic relationships and continuous motion dynamics. Recent SLP works have mainly followed three families: autoregressive models, diffusion models, and flow-based generative models. Autoregressive methods~\citep{saunders2020progressive} generate sign motions sequentially, and their recent tokenizer-LM variants~\citep{yin2024t2s, Zuo_2025_ICCV} encode continuous sign motion into discrete latent representations to leverage language models for sequential dependencies. Diffusion methods~\citep{baltatzis2024neural, tang2025sign, sung2026context} generate sign motions by iterative denoising under text-conditioned guidance. Among them, GARD~\citep{sung2026context} refines inter-gloss transitions with the geodesic distance in place of the Euclidean one, yet the improvement remains local to gloss transitions and its generative transport stays in Euclidean space. Flow matching-based methods~\citep{khan2025signflow, low2026signspark} anchor training on conditional paths but construct these paths by linear interpolation, thereby ignoring the rotational geometry of human joints and the motion distribution of sign motions. These efforts thus center on conditioning and decoding mechanisms, while the geometry of the generative transport remains underexplored. In this work, we construct conditional paths on the motion manifold, jointly considering geometric efficiency and the motion distribution.

\subsection{Flow Matching} 
Training continuous normalizing flows originally required simulating the ODE and backpropagating through the solver, which is computationally expensive. Flow matching~\citep{lipmanflow} and the concurrent stochastic-interpolant formulation~\citep{albergobuilding} remove this cost by regressing a learnable velocity field onto per-pair conditional paths, yielding a simulation-free objective anchored on conditional paths. Subsequent work studies how these paths should be designed. Rectified flow~\citep{liu2022flow} transports samples along the linear interpolation between paired endpoints and iteratively straightens the trajectories to enable few-step sampling, and minibatch optimal transport~\citep{tong2024tmlr-improving} couples the endpoints to reduce path crossing. Flow matching on general geometries~\citep{chen2024flow} further replaces the linear interpolation with geodesic interpolation on Riemannian manifolds, so that conditional paths respect the intrinsic geometry of the data. Flow matching has also entered SLP. SignFlow~\citep{khan2025signflow} introduces conditional flow matching with optimal-transport coupling for text-driven sign motion generation, and SignSparK~\citep{low2026signspark} incorporates sparse keyframes into the flow matching framework to guide generation. Both methods construct their conditional paths by linear interpolation in Euclidean space, treating rotation representations as unconstrained Euclidean variables.

\section{Detailed Derivation of Property~\ref{prop: geometric}}
\label{app: derivation}
In this section, we provide the detailed derivation that completes the proof of Property~\ref{prop: geometric}, and then extend the analysis to the remaining standard rotation representations. We first work directly on the rotation group $SO(3)$ (Sec.~\ref{app: sub-so3}), then show that the 6D rotation representation adopted by existing flow matching-based SLP methods inherits exactly the same
off-manifold deviation (Sec.~\ref{app: sub-6d}), and finally establish the deviation for the quaternion representation (Sec.~\ref{app: sub-quaternion}) and the Euler-angle representation (Sec.~\ref{app: sub-euler}).

\subsection{Derivation on \texorpdfstring{$SO(3)$}{SO(3)}}
\label{app: sub-so3}
\textbf{Setup.}
Recall that $SO(3)=\{Q\in\mathbb{R}^{3\times 3}\mid Q^{\top}Q=I,\ \det(Q)=1\}$, and every $R\in SO(3)$ has orthonormal columns and hence satisfies $\Vert R \Vert_{F}^{2}=\mathrm{tr}(R^{\top}R)=3$. Given two distinct rotations $R_{0},R_{1}\in SO(3)$, the conditional path constructed by linear interpolation (Eq.~\ref{eq: rot-path} in the main text) is
\begin{equation}
R_{t} = (1 - t)\,R_{0} + t\,R_{1}, \quad t\in[0,1].
\end{equation}

\textbf{Derivation.} Expanding $\Vert R_t \Vert_{F}^{2}=\langle R_{t}, R_{t}\rangle_{F}$ using the bilinearity of the Frobenius inner product yields
\begin{align}
\Vert R_{t} \Vert_{F}^{2}
&= (1-t)^{2}\,\Vert R_{0} \Vert_{F}^{2}
   + 2t(1-t)\,\langle R_{0}, R_{1} \rangle_{F}
   + t^{2}\,\Vert R_{1} \Vert_{F}^{2} \notag \\
&= 3\big[(1-t)^{2} + t^{2}\big] + 2t(1-t)\,\langle R_{0}, R_{1} \rangle_{F}.
   \label{eq: app-rot-expand}
\end{align}
On the other hand, the squared Frobenius distance between the two endpoints expands as
\begin{align}
\Vert R_{0} - R_{1} \Vert_{F}^{2}
&= \Vert R_{0} \Vert_{F}^{2} - 2\,\langle R_{0}, R_{1} \rangle_{F}
   + \Vert R_{1} \Vert_{F}^{2} \notag \\
&= 6 - 2\,\langle R_{0}, R_{1} \rangle_{F},
   \label{eq: app-rot-dist}
\end{align}
which gives $\langle R_{0}, R_{1} \rangle_{F} = 3 - \tfrac{1}{2}\Vert R_{0} - R_{1} \Vert_{F}^{2}$. Substituting this relation into Eq.~\ref{eq: app-rot-expand} and using the identity $(1-t)^{2} + t^{2} = 1 - 2t(1-t)$ completes the derivation:
\begin{equation}
\Vert R_{t} \Vert_{F}^{2}
= 3 - t(1-t)\,\Vert R_{0} - R_{1} \Vert_{F}^{2}.
\label{eq: app-rot-final}
\end{equation}
Moreover, writing $\theta$ for the relative rotation angle of $R_{0}^{\top}R_{1}$, the Frobenius inner product between two rotations admits the closed form $\langle R_{0}, R_{1} \rangle_{F} = \mathrm{tr}(R_{0}^{\top}R_{1}) = 1 + 2\cos\theta$, so that $\Vert R_{0} - R_{1} \Vert_{F}^{2} = 4(1-\cos\theta)$ and Eq.~\ref{eq: app-rot-final} reads $\Vert R_{t} \Vert_{F}^{2} = 3 - 4t(1-t)(1-\cos\theta) < 3$ whenever $\theta \neq 0$, i.e., $R_{0}\neq R_{1}$. The determinant shrinks in parallel: the eigenvalues of $R_{0}^{\top}R_{1}$ are $1$ and $e^{\pm i\theta}$, so $\det(R_{t}) = \det\big(R_{0}\big[(1-t)I + t\,R_{0}^{\top}R_{1}\big]\big) = 1 - 2t(1-t)(1-\cos\theta) < 1$, and the orthogonality and determinant constraints of $SO(3)$ are violated simultaneously.

\textbf{Completion of Property~\ref{prop: geometric}.}
If $R_{0}\neq R_{1}$, then $\Vert R_{0}-R_{1}\Vert_{F}^{2} > 0$. For any $t\in(0,1)$ we have $t(1-t)>0$, and Eq.~\ref{eq: app-rot-final} yields
\begin{equation}
\Vert R_{t} \Vert_{F}^{2}
= 3 - t(1-t)\,\Vert R_{0} - R_{1} \Vert_{F}^{2} < 3.
\end{equation}
Since every $R\in SO(3)$ satisfies $\Vert R \Vert_{F}^{2}=3$, the interpolated state $R_{t}$ violates the orthogonality constraint $R_{t}^{\top}R_{t}=I$, and therefore $R_{t}\notin SO(3)$ for all $t\in(0,1)$.

\subsection{Derivation on the 6D Rotation Representation}
\label{app: sub-6d}
In practice, existing flow matching-based SLP methods~\citep{khan2025signflow, low2026signspark} represent joint rotations with the 6D rotation representation~\citep{zhou2019continuity} $\psi(R) = (a, b)$, which retains the first two columns of $R$. Since $\psi$ is a linear map, $\psi(R_{t}) = (1-t)\,\psi(R_{0}) + t\,\psi(R_{1})$, so the off-manifold deviation established in Sec.~\ref{app: sub-so3} persists on the embedded 6D manifold. We make this precise at the coordinate level below.

\textbf{Setup.} The manifold of valid 6D representations is $\mathcal{M}_{\mathrm{6D}}=\{(a,b)\in\mathbb{R}^{3}\times\mathbb{R}^{3} \mid \Vert a \Vert_{2}=\Vert b \Vert_{2}=1,\ a^{\top}b=0\}$. Given two points $x_{0}=(a_{0},b_{0})$ and $x_{1}=(a_{1},b_{1})$ on
$\mathcal{M}_{\mathrm{6D}}$, the conditional path constructed by linear interpolation is
\begin{equation}
x_{t} = (1-t)\,x_{0} + t\,x_{1} = (a_{t},\, b_{t}), \quad t\in[0,1],
\end{equation}
where $a_{t}=(1-t)a_{0} + ta_{1}$ and $b_{t}=(1-t)b_{0} + tb_{1}$. Since $x_{0}, x_{1}\in\mathcal{M}_{\mathrm{6D}}$, their components satisfy the unit-norm constraints
\begin{equation}
\Vert a_{0}\Vert_{2}^{2} = \Vert a_{1}\Vert_{2}^{2}
= \Vert b_{0}\Vert_{2}^{2} = \Vert b_{1}\Vert_{2}^{2} = 1.
\end{equation}

\textbf{Derivation.} Expanding the squared norm of $a_{t}$ via the bilinearity of the inner product yields
\begin{align}
\Vert a_{t} \Vert_{2}^{2}
&= \big\Vert (1 - t)a_{0} + ta_{1} \big\Vert^{2}_{2} \notag \\
&= \big\langle (1-t)a_{0} + ta_{1},\; (1-t)a_{0} + ta_{1} \big\rangle \notag \\
&= (1-t)^{2}\,\Vert a_{0} \Vert_{2}^{2}
   + 2t(1-t)\, a_{0}^{\top}a_{1}
   + t^{2}\,\Vert a_{1} \Vert_{2}^{2}.
   \label{eq: app-expand}
\end{align}
Substituting the unit-norm constraints $\Vert a_{0} \Vert_{2}^{2} = \Vert a_{1} \Vert_{2}^{2} = 1$ into Eq.~\ref{eq: app-expand} and using the identity $(1-t)^{2} + t^{2} = 1 - 2t(1-t)$, we obtain
\begin{align}
\Vert a_{t} \Vert_{2}^{2}
&= (1-t)^{2} + t^{2} + 2t(1-t)\, a_{0}^{\top}a_{1} \notag \\
&= 1 - 2t(1-t)\big(1 - a_{0}^{\top}a_{1}\big).
   \label{eq: app-middle}
\end{align}
On the other hand, the squared distance between the two endpoints expands as
\begin{align}
\Vert a_{0} - a_{1} \Vert_{2}^{2}
&= \Vert a_{0} \Vert_{2}^{2} - 2\,a_{0}^{\top}a_{1} + \Vert a_{1} \Vert_{2}^{2}
\notag \\
&= 2\big(1 - a_{0}^{\top}a_{1}\big),
   \label{eq: app-dist}
\end{align}
which gives $1 - a_{0}^{\top}a_{1} = \tfrac{1}{2}\Vert a_{0} - a_{1} \Vert_{2}^{2}$. Substituting this relation into Eq.~\ref{eq: app-middle} completes the derivation:
\begin{equation}
\Vert a_{t} \Vert_{2}^{2}
= \big\Vert (1 - t)a_{0} + ta_{1} \big\Vert^{2}_{2}
= 1 - t(1-t)\,\Vert a_{0} - a_{1} \Vert_{2}^{2}.
\label{eq: app-final}
\end{equation}

\textbf{Completion.} If $a_{0}\neq a_{1}$, then by the Cauchy--Schwarz inequality $a_{0}^{\top}a_{1} \le \Vert a_{0}\Vert_{2}\Vert a_{1}\Vert_{2} = 1$, with equality if and only if $a_{0}=a_{1}$; hence $\Vert a_{0}- a_{1}\Vert_{2}^{2} = 2(1 - a_{0}^{\top}a_{1}) > 0$ by Eq.~\ref{eq: app-dist}. For any $t\in(0,1)$ we have $t(1-t)>0$, and Eq.~\ref{eq: app-final} yields
\begin{equation}
\Vert a_{t} \Vert_{2}^{2} = 1 - t(1-t)\,\Vert a_{0} - a_{1} \Vert_{2}^{2} < 1,
\end{equation}
so $a_{t}$ violates the unit-norm constraint. If instead $a_{0}=a_{1}$ but $x_{0}\neq x_{1}$, then $b_{0}\neq b_{1}$, and the same argument applied to the $b$ component shows $\Vert b_{t} \Vert_{2}^{2} < 1$. In either case, at least one component of $x_{t}$ violates the defining constraints of $\mathcal{M}_{\mathrm{6D}}$, and therefore $x_{t}\notin\mathcal{M}_{\mathrm{6D}}$ for all $t\in(0,1)$, in exact correspondence with the matrix-level deviation established in Sec.~\ref{app: sub-so3}.

\subsection{Derivation on the Quaternion Representation}
\label{app: sub-quaternion}
Unit quaternions are the most compact minimal parameterization of rotations: each rotation corresponds to a point of $\mathbb{S}^{3}=\{q\in\mathbb{R}^{4}\mid\Vert q\Vert_{2}=1\}$, and $\mathbb{S}^{3}$ is the representation adopted by SignNMFlow (Sec.~\ref{sec: representation}). We show that the conditional path constructed by linear interpolation between unit quaternions leaves $\mathbb{S}^{3}$ through exactly the same norm-contraction mechanism as on $SO(3)$ and the 6D representation.

\textbf{Setup.} Given two distinct unit quaternions $q_{0},q_{1}\in\mathbb{S}^{3}$, the conditional path constructed by linear interpolation is
\begin{equation}
q_{t}=(1-t)\,q_{0}+t\,q_{1},\qquad t\in[0,1],
\end{equation}
where both endpoints satisfy the unit-norm constraint $\Vert q_{0}\Vert_{2}^{2}=\Vert q_{1}\Vert_{2}^{2}=1$.

\textbf{Derivation.} Expanding the squared norm of $q_{t}$ via the bilinearity of the inner product yields
\begin{align}
\Vert q_{t}\Vert_{2}^{2}
&=(1-t)^{2}\,\Vert q_{0}\Vert_{2}^{2}
+2t(1-t)\,q_{0}^{\top}q_{1}
+t^{2}\,\Vert q_{1}\Vert_{2}^{2}\notag\\
&=1-2t(1-t)\big(1-q_{0}^{\top}q_{1}\big).
\label{eq: app-quat-middle}
\end{align}
On the other hand, the squared distance between the two endpoints expands as
\begin{align}
\Vert q_{0}-q_{1}\Vert_{2}^{2}
&=\Vert q_{0}\Vert_{2}^{2}-2\,q_{0}^{\top}q_{1}+\Vert q_{1}\Vert_{2}^{2}\notag\\
&=2\big(1-q_{0}^{\top}q_{1}\big),
\label{eq: app-quat-dist}
\end{align}
which gives $1-q_{0}^{\top}q_{1}=\tfrac{1}{2}\Vert q_{0}-q_{1}\Vert_{2}^{2}$. Substituting this relation into Eq.~\ref{eq: app-quat-middle} completes the derivation:
\begin{equation}
\Vert q_{t}\Vert_{2}^{2}
=\big\Vert(1-t)\,q_{0}+t\,q_{1}\big\Vert_{2}^{2}
=1-t(1-t)\,\Vert q_{0}-q_{1}\Vert_{2}^{2}.
\label{eq: app-quat-final}
\end{equation}
Writing $\theta=\arccos(q_{0}^{\top}q_{1})\in[0,\pi]$ for the spherical distance between the two representatives, Eq.~\ref{eq: app-quat-dist} gives $\Vert q_{0}-q_{1}\Vert_{2}^{2}=2(1-\cos\theta)=4\sin^{2}(\theta/2)$, so the deviation also reads $\Vert q_{t}\Vert_{2}^{2}=1-4t(1-t)\sin^{2}(\theta/2)$, in exact correspondence with the angle form $\Vert R_{t}\Vert_{F}^{2}=3-4t(1-t)(1-\cos\theta)$ of Sec.~\ref{app: sub-so3}, with $\theta$ playing the role of the half rotation angle.

\textbf{Completion.} If $q_{0}\neq q_{1}$, then by the Cauchy--Schwarz inequality $q_{0}^{\top}q_{1}\le\Vert q_{0}\Vert_{2}\Vert q_{1}\Vert_{2}=1$, with equality if and only if $q_{0}=q_{1}$; hence $\Vert q_{0}-q_{1}\Vert_{2}^{2}>0$ by Eq.~\ref{eq: app-quat-dist}. For any $t\in(0,1)$ we have $t(1-t)>0$, and Eq.~\ref{eq: app-quat-final} yields
\begin{equation}
\Vert q_{t}\Vert_{2}^{2}=1-t(1-t)\,\Vert q_{0}-q_{1}\Vert_{2}^{2}<1,
\end{equation}
so $q_{t}$ violates the unit-norm constraint that defines $\mathbb{S}^{3}$, and therefore $q_{t}\notin\mathbb{S}^{3}$ for all $t\in(0,1)$. On the motion manifold $\mathcal{M}=(\mathbb{S}^{3})^{J}$ of Sec.~\ref{sec: representation}, linear interpolation between two motions applies this computation joint by joint, so every joint of every interior state leaves its hypersphere factor. The deviation is thus present in the raw quaternion coordinates as well, prior to any manifold structure being imposed, and it is again the Exp-based construction of Sec.~\ref{sec: path} (Appendix~\ref{app: exactness}) that restores manifold validity.

\textbf{Remark (renormalized interpolation).} 
A natural remedy is to project the interpolated state back onto the sphere, $\hat{q}_{t}=q_{t}/\Vert q_{t}\Vert_{2}$, which yields normalized linear interpolation (nlerp). In the two-plane $\mathrm{span}\{q_{0},q_{1}\}$, write $q_{0}=e_{1}$ and $q_{1}=\cos\theta\,e_{1}+\sin\theta\,e_{2}$ with $\theta\in(0,\pi)$; the renormalized state $\hat{q}_{t}$ traces the same great-circle arc as the spherical interpolation (slerp), but at the state-dependent arc angle $\chi(t)=\mathrm{atan2}\big(t\sin\theta,\ 1-t+t\cos\theta\big)$, and differentiating gives $\dot{\chi}(t)=\sin\theta/\Vert q_{t}\Vert_{2}^{2}$. Since $\Vert q_{t}\Vert_{2}^{2}$ attains its minimum at $t=\tfrac{1}{2}$, the angular speed peaks mid-path and exceeds its endpoint value by the factor
\begin{equation}
\frac{\dot{\chi}(\tfrac{1}{2})}{\dot{\chi}(0)}
=\frac{1}{\Vert q_{1/2}\Vert_{2}^{2}}
=\frac{2}{1+\cos\theta}
=\sec^{2}\!\big(\tfrac{\theta}{2}\big),
\label{eq: app-quat-nlerp}
\end{equation}
so the renormalized path departs from the constant-speed geodesic whenever$\theta\neq0$, and the distortion grows without bound as the representatives approach antipodes ($\theta\to\pi$). Renormalization therefore restores the constraint but not geometric efficiency: the states sweep the connecting arc at a state-dependent speed, in contrast to the constant-speed geodesic reference that Sec.~\ref{sec: path} adopts.

\subsection{Derivation on the Euler-Angle Representation}
\label{app: sub-euler}
Euler angles parameterize a rotation by three angles $u=(\alpha,\beta,\gamma)$ through a fixed axis factorization. We take the ZYX convention $R(u)=R_{z}(\alpha)\,R_{y}(\beta)\,R_{x}(\gamma)$, where $R_{z},R_{y},R_{x}$ denote rotations about the axes of the standard Euclidean frame (the other conventions are analogous). Unlike the representations above, the raw parameter space $\mathbb{R}^{3}$ carries no algebraic constraint: every angle triple parameterizes a valid rotation. The incompatibility between linear
interpolation and rotational geometry therefore manifests through two complementary mechanisms, which we derive in turn: violating the periodic structure of the faithful parameter manifold, and distorting the induced rotation path when the raw angles are treated as unconstrained Euclidean variables.

\textbf{Derivation I: periodicity mismatch on the faithful parameter manifold.} Each Euler angle is defined modulo $2\pi$, so the faithful parameter manifold is the torus $\mathbb{T}^{3}=(\mathbb{S}^{1})^{3}$, embedded in $\mathbb{R}^{6}$ through the standard circle embedding $\eta(\alpha)=(\cos\alpha,\sin\alpha)\in\mathbb{S}^{1}\subset\mathbb{R}^{2}$ applied per angle. Linear interpolation between two embedded representations $x_{0},x_{1}\in\mathbb{T}^{3}$ is the per-coordinate interpolation $x_{t}=(1-t)x_{0}+t\,x_{1}$ in $\mathbb{R}^{6}$. Fixing one angle and writing $\Delta=\alpha_{1}-\alpha_{0}$ for its increment, the bilinearity of the inner product gives
\begin{align}
\big\Vert(1-t)\,\eta(\alpha_{0})+t\,\eta(\alpha_{1})\big\Vert_{2}^{2}
&=(1-t)^{2}+t^{2}+2t(1-t)\cos\Delta\notag\\
&=1-2t(1-t)\big(1-\cos\Delta\big)\notag\\
&=1-t(1-t)\,\big\Vert\eta(\alpha_{0})-\eta(\alpha_{1})\big\Vert_{2}^{2},
\label{eq: app-euler-torus}
\end{align}
using $\Vert\eta(\alpha_{0})-\eta(\alpha_{1})\Vert_{2}^{2}=2-2\cos\Delta=4\sin^{2}(\Delta/2)>0$ whenever $\alpha_{0}\not\equiv\alpha_{1}\ (\mathrm{mod}\ 2\pi)$. Each circle factor of the interpolated state therefore violates its unit-norm constraint for every $t\in(0,1)$, with the deviation $4t(1-t)\sin^{2}(\Delta/2)$ maximized, over the increment, when the two angles differ by an odd multiple of $\pi$. This is the norm-contraction of Secs.~\ref{app: sub-so3}--\ref{app: sub-quaternion} instantiated on $\mathbb{S}^{1}$. The conditional path constructed by linear interpolation leaves the embedded torus $\mathbb{T}^{3}$ whenever the two representations differ.

\textbf{Derivation II: distortion of raw-angle interpolation.} In practice, Euler angles would be stored and interpolated as unconstrained Euclidean vectors, $u_{t}=(1-t)\,u_{0}+t\,u_{1}\in\mathbb{R}^{3}$, the Euler analogue of treating rotation coordinates as flat Euclidean variables (Sec.~\ref{sec: representation}). Every intermediate triple is a valid parameterization, so no intermediate state is formally off-manifold; the failure instead appears in the geometry of the induced rotation path $t\mapsto R(u_{t})$, in two quantifiable ways.

First, the transport is state-dependent. Let $\omega_{b}(t)\in\mathbb{R}^{3}$ denote the angular velocity of $R(u_{t})$ expressed in the moving frame, i.e., the unique vector satisfying $\dot{R}(u_{t})=R(u_{t})\,[\omega_{b}(t)]_{\times}$, where $[\cdot]_{\times}$ is the cross-product matrix. Differentiating the factorization gives
\begin{equation}
\omega_{b}(t)=\dot{\gamma}\,\hat{x}+\dot{\beta}\,R_{x}(\gamma)^{\top}\hat{y}+\dot{\alpha}\,R_{x}(\gamma)^{\top}R_{y}(\beta)^{\top}\hat{z}=J(u_{t})\,(u_{1}-u_{0}),
\label{eq: app-euler-omega}
\end{equation}
with the constant rate $\dot{u}_{t}=u_{1}-u_{0}$ and the state-dependent Jacobian
\begin{equation}
J(u)=
\begin{pmatrix}
-\sin\beta & 0 & 1\\
\cos\beta\,\sin\gamma & \cos\gamma & 0\\
\cos\beta\,\cos\gamma & -\sin\gamma & 0
\end{pmatrix},
\qquad \det J(u)=-\cos\beta,
\label{eq: app-euler-jac}
\end{equation}
whose determinant vanishes exactly on the gimbal-lock locus $\{\beta=\tfrac{\pi}{2}+k\pi,\ k\in\mathbb{Z}\}$. Along the geodesic, the rotation proceeds at constant speed about a fixed axis; along the Euler-interpolated path, both the direction and the magnitude of $\omega_{b}(t)$ vary with $t$ through $u_{t}$. At gimbal lock, $J$ loses rank and the parameterization fails to be a local diffeomorphism: since $\sigma_{\min}\big(J(u)\big)\le\vert\det J(u)\vert^{1/3}\to0$ as $\beta\to\tfrac{\pi}{2}+k\pi$ (while $\Vert J(u)\Vert_{F}=\sqrt{3}$ for every $u$, so the collapse is purely anisotropic), the inverse map degenerates, $\Vert J(u)^{-1}\Vert_{2}=1/\sigma_{\min}\big(J(u)\big)\to\infty$. Bounded Euclidean increments of the angles thus map to collapsing rotation increments in the near-kernel direction, while the angle rates required to realize angular velocities in the corresponding output direction diverge. Euclidean increments in parameter space carry no consistent rotation-geometric meaning near this locus.

Second, the parameterization is non-injective, and Euclidean interpolation fabricates spurious excursions between representations of the same rotation. Consider $u_{0}=(0,0,0)$ and $u_{1}=(\pi,\pi,\pi)$: the half-turn rotations $R_{z}(\pi)$, $R_{y}(\pi)$, $R_{x}(\pi)$ are diagonal with diagonal entries $(-1,-1,1)$, $(-1,1,-1)$, $(1,-1,-1)$, whose product is the identity, so both triples parameterize $R=I$ and the geodesic transport between them is the constant path, with geodesic distance $d\big(R(u_{0}),R(u_{1})\big)=0$. The linearly interpolated path $u_{t}=t\,(\pi,\pi,\pi)$ is nevertheless non-constant, and its midpoint evaluates to
\begin{equation}
R\big(\tfrac{\pi}{2},\tfrac{\pi}{2},\tfrac{\pi}{2}\big)
=
\begin{pmatrix}
0 & 0 & 1\\
0 & 1 & 0\\
-1 & 0 & 0
\end{pmatrix}
=R_{y}\big(\tfrac{\pi}{2}\big),
\label{eq: app-euler-mid}
\end{equation}
a rotation of angle $\tfrac{\pi}{2}$ about the $y$-axis. Euclidean interpolation in angle space therefore produces an intermediate state at angular distance at least $\tfrac{\pi}{2}$ from both endpoints,
\begin{equation}
\sup_{t\in[0,1]}\,d\big(R(u_{t}),\,R(u_{0})\big)\ \ge\ \tfrac{\pi}{2}\ >\ 0\ =\ d\big(R(u_{0}),\,R(u_{1})\big),
\label{eq: app-euler-sup}
\end{equation}
where $d(\cdot,\cdot)$ denotes the geodesic distance on $SO(3)$ (the rotation angle of the relative rotation). The spurious excursion is an instance of the standard aliasing identity $R(\alpha{+}\pi,\ \pi{-}\beta,\ \gamma{+}\pi)=R(\alpha,\beta,\gamma)$, under which $u_{1}$ is an alternative parameterization of $u_{0}$: because the parameterization folds $\mathbb{R}^{3}$ onto $SO(3)$, linear structure in parameter space carries no rotation-geometric meaning.

\textbf{Completion.} For the Euler-angle representation, the faithful parameter manifold $\mathbb{T}^{3}$ suffers the same off-manifold deviation as $SO(3)$ (Sec.~\ref{app: sub-so3}), the 6D representation (Sec.~\ref{app: sub-6d}), and the quaternion representation (Sec.~\ref{app: sub-quaternion}) (Derivation I), while interpolating the constraint-free raw angles substitutes a complementary failure: a state-dependent, rank-deficient, and non-injective distortion of the induced rotation path (Derivation II). Both mechanisms stem from applying Euclidean-linear operations to rotation parameters. Across the four standard rotation representations, linear interpolation therefore either leaves the representation manifold or distorts the rotation path itself; constructing conditional paths on $\mathcal{M}$ through the exponential map (Sec.~\ref{sec: path}) avoids both failure modes.

\section{Additional Details for Section~\ref{sec: nmf}}
\subsection{Quaternion Sign Convention}
\label{app: quaternion-sign}
Unit quaternions $q$ and $-q$ represent the same rotation. Under the scalar-first convention $(q_w, q_x, q_y, q_z)$, we resolve this sign ambiguity by selecting the representative whose first nonzero component is positive. For each quaternion $\tilde q_{j}=(\tilde q_{j,w},\tilde q_{j,x},\tilde q_{j,y},\tilde q_{j,z})$ of the motion $x$ in Eq.~\ref{eq: representation}, the sign selection is
\begin{equation}
    q_{j}=
    \begin{cases}
        -\tilde q_{j}, & \tilde q_{j,w}<0,\\
        \phantom{-}\tilde q_{j}, & \tilde q_{j,w}> 0,\\
        \phantom{-}\sigma_{j}\tilde q_{j}, & \tilde q_{j,w}= 0,
    \end{cases}
    \label{eq: quaternion-sign}
\end{equation}
where $\tilde q_{j}$ is the quaternion before sign adjustment, $q_{j}$ is the selected representative, and $\sigma_{j}\in\{+1,-1\}$ makes the first nonzero component of $(\tilde q_{j,x},\tilde q_{j,y},\tilde q_{j,z})$ positive. This selects a unique representative for every rotation, including the $180^\circ$ case. All representatives lie in a closed hemisphere of $\mathbb{S}^{3}$, and no two of them are antipodal, so the logarithmic maps in Eq.~\ref{eq: explog} are always evaluated between non-antipodal quaternions.

\subsection{Prototype Extraction}
\label{app: sub-prototype}
Given a training set containing $N$ motion frames, we extract prototypical motions to characterize the motion distribution. For each joint independently, we perform K-means clustering~\citep{mcqueen1967some} over all joint rotations and obtain $K$ prototypes. The process can be formalized as:
\begin{equation}
    \lbrace \tilde{c}^{j}_{k}\rbrace^{K}_{k=1}=\mathop{\arg\min}_{\{\mu_k\}_{k=1}^K}
    \sum_{n=1}^N \mathop{\min}_{k \in \{1, \dots, K\}} \left\| q_n^j - \mu_k
    \right\|_2^2,
    \quad c^{j}_{k}=\frac{\tilde{c}^{j}_{k}}{\Vert\tilde{c}^{j}_{k}\Vert_{2}},
    \quad k = 1, \dots, K,
\end{equation}
where $q_n^j$ denotes the sign-aligned quaternion in Eq.~\ref{eq: quaternion-sign} of joint $j$ at frame $n$, and $c^{j}_{k}$ is the normalized prototype. To preserve the occurrence information of the motions, we further associate each prototype with an empirical weight:
\begin{equation}
    w^{j}_{k} = \frac{|\mathcal{I}_k^j|}{N},
    \qquad
    \sum_{k=1}^{K} w^{j}_{k} = 1,
\end{equation}
where $\mathcal{I}_k^j$ denotes the set of rotations assigned to the $k$-th cluster of the $j$-th joint, and $w^{j}_{k}$ represents the corresponding empirical occurrence probability. The prototypes $\{c_k\}_{k=1}^{K}$ and frequency weights $\{w_k\}_{k=1}^{K}$ in Sec.~\ref{sec: measure} are instantiated joint-wise as $c_k = c^{j}_{k}$ and $w_k = w^{j}_{k}$ within each factor of the product manifold. The renormalized centroid $c^{j}_{k}$ coincides with the spherical K-means centroid and closely approximates the Riemannian (Fréchet) mean whenever the cluster is concentrated.

\subsection{Validity of the Motion Naturalness Measure}
\label{app: sub-validity}
We establish the validity of the measure through four analyses: $G$ is well defined on $\mathcal{M}$, the density factor consistently estimates the motion distribution, the induced transport cost is inversely weighted by motion density, and the measure degenerates to the standard geometry under uniform distribution.

\textbf{Setup and notation.} The naturalness measure reads $G(x) = m(x)\big(S(x) + \rho\mathbf{I}\big)^{-1}$, where $m(x)$ denotes the aggregate kernel mass. Recall that $\delta_k = \mathrm{Log}_{x}(c_k)\in T_{x}\mathcal{M}$ with $\|\delta_k\| = d_{\mathcal{M}}(x, c_k)$, the geodesic distance on $\mathcal{M}$, and that $T_{x}\mathcal{M}\cong\mathbb{R}^{3J}$ carries the Euclidean inner product. Each prototype retains a non-empty cluster, so its frequency weight satisfies $w_k > 0$.

\textbf{Well-definedness.} The naturalness norm (Eq.~\ref{eq: path-naturalness}) and the kinetic energy objective (Eq.~\ref{eq: ln}) require $G(x)$ to be invertible at every point of the manifold, and the regularized construction in Eq.~\ref{eq: naturalness} guarantees this. Since $w_k>0$ and the Gaussian kernel is strictly positive, $\tilde{w}_k(x)>0$ for all $k$ and all $x$, hence $m(x)>0$. Each outer product $\delta_k\delta_k^{\top}$ is symmetric positive semidefinite (PSD), and a nonnegative combination of PSD matrices is PSD, so $S(x)\succeq 0$ and therefore $S(x)+\rho\mathbf{I}\succeq\rho\mathbf{I}\succ 0$. The inverse of a symmetric positive definite (SPD) matrix is SPD, and a positive multiple of an SPD matrix is SPD, so $G(x)=m(x)\big(S(x)+\rho\mathbf{I}\big)^{-1}\succ 0$ at every $x\in\mathcal{M}$. Consequently, $G(x)^{-1}$ exists on $\mathcal{M}$, and $\mathcal{N}(\gamma)$ (Eq.~\ref{eq: path-naturalness}), $\varphi(\gamma)$ (Eq.~\ref{eq: psi}), and $\mathcal{L}_{G}(\theta)$ (Eq.~\ref{eq: ln}) are well-defined. Moreover, $G(x)^{-1}=\frac{1}{m(x)}\big(S(x)+\rho\mathbf{I}\big)$ has eigenvalues in $\big[\rho/m(x),\,(\lambda_{\max}(S(x))+\rho)/m(x)\big]$, and the Rayleigh quotient of a symmetric matrix lies between its extreme eigenvalues, so for every $v\in T_{x}\mathcal{M}$,
\begin{equation}
    \frac{\rho}{m(x)}\,\|v\|^{2}
    \;\le\;
    v^{\top}G(x)^{-1}v
    \;\le\;
    \frac{\lambda_{\max}\big(S(x)\big)+\rho}{m(x)}\,\|v\|^{2},
    \label{eq: app-bounds}
\end{equation}
where $\lambda_{\max}(\cdot)$ denotes the largest eigenvalue. These bounds quantify how the regularization constant $\rho$ and the aggregate density $m(x)$ jointly control the transport cost.

\textbf{Density consistency.} 
We show that the density factor $m(x)$ is a kernel density estimate (KDE) of the motion distribution on $\mathcal{M}$, compressed onto the $K$ prototypes with error controlled by $K$. Let $\{x_i\}_{i=1}^{N}$ be the training motions. Let $\mathcal{C}_k$ be the cluster assigned to prototype $c_k$, with frequency weight $w_k = |\mathcal{C}_k|/N$, and let $\kappa_\sigma(r)=\exp(-r^{2}/2\sigma^{2})$. Eq.~\ref{eq: naturalness} then gives $m(x)=\frac{1}{N}\sum_{k=1}^{K}|\mathcal{C}_k|\, \kappa_\sigma\big(d_{\mathcal{M}}(x,c_k)\big)$. Let $\hat{p}_\sigma(x)=\frac{1}{N}\sum_{i=1}^{N}\kappa_\sigma\big(d_{\mathcal{M}}(x,x_i)\big)$ be the KDE built directly from the training motions. The geodesic distance satisfies the triangle inequality, so $\big|d_{\mathcal{M}}(x,x_i)-d_{\mathcal{M}}(x,c_{k(i)})\big|\le d_{\mathcal{M}}(x_i,c_{k(i)})$, where $k(i)$ indexes the cluster containing $x_i$. The kernel $\kappa_\sigma$ is globally Lipschitz with constant $L_\sigma=\max_{r\ge 0}|\kappa_\sigma'(r)|=\frac{1}{\sigma\sqrt{e}}$, attained at $r=\sigma$, so each summand obeys $\big|\kappa_\sigma(d_{\mathcal{M}}(x,x_i))-\kappa_\sigma(d_{\mathcal{M}}(x, c_{k(i)}))\big|\le\frac{1}{\sigma\sqrt{e}}\,d_{\mathcal{M}}(x_i,c_{k(i)})$, and averaging over $i$ yields
\begin{equation}
    \big|m(x)-\hat{p}_\sigma(x)\big|
    \;\le\;
    \frac{\bar{r}_K}{\sigma\sqrt{e}},
    \qquad
    \bar{r}_K := \frac{1}{N}\sum_{i=1}^{N}d_{\mathcal{M}}\big(x_i,\,c_{k(i)}\big).
    \label{eq: app-quantization}
\end{equation}
Moreover, the optimal quantization error $\bar{r}_K^{\star}$, the minimum of $\bar{r}_K$ over all $K$-prototype configurations, is non-increasing in $K$: any optimal $K$-prototype configuration induces a feasible $(K{+}1)$-prototype configuration of identical cost by duplicating one prototype, so $\bar{r}_{K+1}^{\star}\le\bar{r}_K^{\star}$. That $\bar{r}_K^{\star}\to 0$ as $K\to\infty$ for data on a compact manifold is a standard result in quantization theory~\citep{pollard1981strong}.

The kernel $\kappa_\sigma$ depends on the pair $(x,c)$ only through the geodesic distance, replacing it with the heat kernel profile, which on the homogeneous manifold $\mathcal{M}$ is likewise a fixed radial function of $d_{\mathcal{M}}$. Replacing $\kappa_\sigma$ with this profile therefore turns $m(x)$ into a standard KDE on $\mathcal{M}$. Its consistency with the population motion density as $N\to\infty$ and $\sigma\to 0$ follows from the classical theory of kernel density estimation on Riemannian manifolds~\citep{pelletier2005kernel}. The density factor of $G$ is therefore a consistent estimate of the motion distribution. Compression onto prototypes adds the error $\bar{r}_K$, which is controlled by $K$ (Eq.~\ref{eq: app-quantization}).

\textbf{Distribution-aware transport cost.} We next show how the measure injects the motion distribution into the transport cost. Expanding the quadratic form of $S(x)$ (Eq.~\ref{eq: scatter}) gives $v^{\top}S(x)v=\sum_{k=1}^{K}\tilde{w}_k(x)\,\langle\delta_k,v\rangle^{2}$, and substituting this into the identity $G(x)^{-1}=\frac{1}{m(x)}\big(S(x)+\rho\mathbf{I}\big)$ from the well-definedness analysis yields, for every $x\in\mathcal{M}$ and $v\in T_{x}\mathcal{M}$,
\begin{equation}
    v^{\top}G(x)^{-1}v
    =
    \frac{1}{m(x)}
    \Big(
    \rho\,\|v\|^{2}
    +
    \sum_{k=1}^{K}\tilde{w}_k(x)\,\big\langle\delta_k,\,v\big\rangle^{2}
    \Big),
    \label{eq: app-cost}
\end{equation}
whose numerator is a sum of nonnegative terms. The density estimate thus enters the instantaneous transport cost $\|\dot{\gamma}_t\|_{G(\gamma_t)}^{2}$ of Eq.~\ref{eq: path-naturalness} only through the prefactor $1/m(x)$: at a fixed numerator, transport through dense regions of the motion distribution contributes less to $\mathcal{N}(\gamma)$ and $\varphi(\gamma)$.

Equivalently, define the kernel-weighted covariance of the prototypes around $x$,
\begin{equation}
    \Sigma(x) := \frac{S(x)}{m(x)} = \sum_{k=1}^{K}\frac{\tilde{w}_k(x)}{m(x)}
    \,\delta_k\delta_k^{\top},
    \label{eq: app-cov}
\end{equation}
and factor $S(x)=m(x)\Sigma(x)$ out of Eq.~\ref{eq: naturalness}:
\begin{equation}
    G(x) = \Big(\Sigma(x) + \frac{\rho}{m(x)}\mathbf{I}\Big)^{-1}.
    \label{eq: app-mahalanobis}
\end{equation}
This form makes the semantics of naturalness explicit. The metric $G(x)$ is large near the core of a dense, tightly concentrated cluster (large $m(x)$, small $\Sigma(x)$) and small in sparse regions or locally ambiguous zones where the prototypes around $x$ disperse. The transport cost in Eq.~\ref{eq: app-cost} is governed by $G(x)^{-1}=\Sigma(x)+\frac{\rho}{m(x)}\mathbf{I}$. It approaches the isotropic form $\frac{\rho}{m(x)}\|\cdot\|^{2}$ within tight cluster cores, while transit through sparse or dispersed regions is expensive. The construction parallels density-normalized diffusion geometry, which reweights the metric by the inverse data density so that transport preferentially follows the data~\citep{coifman2006diffusion}. $\mathcal{L}_{G}$ in Eq.~\ref{eq: ln} therefore lowers the transport cost in regions where sign motions distribute, incorporating the motion distribution into the learned paths.

\paragraph{Geodesic recovery under a uniform distribution.} Finally, we formalize the claim in Sec.~\ref{sec: path} that the naturalness measure degenerates to the standard geometry when the motion distribution is uniform. The learned correction in Eq.~\ref{eq: learnable-path} is then driven solely by the non-uniformity of the distribution. Let $\mu$ be the uniform probability measure on $\mathcal{M}$ (the normalized Riemannian volume), and define the population counterparts of $m(x)$ (Eq.~\ref{eq: naturalness}) and the scatter matrix (Eq.~\ref{eq: scatter}):
\begin{equation}
\begin{aligned}
    m_\mu(x) &= \int_{\mathcal{M}}\kappa_\sigma\big(d_{\mathcal{M}}(x,c)\big)
    \,d\mu(c),\\
    S_\mu(x) &= \int_{\mathcal{M}}\kappa_\sigma\big(d_{\mathcal{M}}(x,c)\big)
    \,\mathrm{Log}_{x}(c)\,\mathrm{Log}_{x}(c)^{\top}\,d\mu(c),
\end{aligned}
\label{eq: app-population}
\end{equation}
where the integrals exclude the $\mu$-null cut locus of $x$. In this regime, the population measure $G_\mu(x):=m_\mu(x)\big(S_\mu(x)+\rho\mathbf{I}\big)^{-1}$ is a constant multiple of the identity. $\varphi(\gamma)$ (Eq.~\ref{eq: psi}) becomes positively proportional to the standard path energy $\int_0^1\|\dot{\gamma}_t\|^{2}\,dt$. Its minimizer over absolutely continuous paths with $\gamma_0=x_0$, $\gamma_1=x_1$ is, up to reparameterization, the constant-speed geodesic $\gamma_t=\mathrm{Exp}_{x_0}\big(t\,\mathrm{Log}_{x_0}(x_1)\big)$, unique whenever $x_1$ does not lie in the cut locus of $x_0$.

To see that $G_\mu$ is constant and isotropic, note that the uniform measure $\mu$ is invariant under every isometry $g$ of the round product manifold $\mathcal{M}$, isometries preserve geodesic distances, and they commute with the logarithmic map, $\mathrm{Log}_{gx}(gc)=g_{*}\,\mathrm{Log}_{x}(c)$. The change of variables $c\mapsto g^{-1}c$ in Eq.~\ref{eq: app-population} therefore gives $m_\mu(gx)=m_\mu(x)$ and $S_\mu(gx)=g_{*}\,S_\mu(x)\,g_{*}^{\top}$. The isometry group of $\mathcal{M}=(\mathbb{S}^{3})^{J}$ acts transitively, so $m_\mu\equiv m_0$ is constant. Taking $g$ in the stabilizer of $x$, which acts on $T_{x}\mathcal{M}=\bigoplus_{j=1}^{J}T_{x^{j}}\mathbb{S}^{3}\cong\mathbb{R}^{3J}$ as arbitrary independent rotations of the $J$ three-dimensional factors together with factor permutations, the identity $S_\mu(x)=g_{*}S_\mu(x)g_{*}^{\top}$ forces $S_\mu(x)=\alpha\mathbf{I}$: blockwise rotation invariance makes $S_\mu(x)$ block-diagonal with scalar blocks, and permutation invariance equalizes the blocks. Transitivity then makes $\alpha$ independent of $x$, and $G_\mu\equiv\frac{m_0}{\alpha+\rho}\mathbf{I}=:c_g\mathbf{I}$.

To identify the energy minimizer, write the expectation over $t\sim\mathcal{U}[0,1]$ as an integral, so that $\varphi(\gamma)=c_g^{-1}\int_0^1\|\dot{\gamma}_t\|^{2}\,dt$. For any absolutely continuous $\gamma$ joining $x_0$ to $x_1$, the Cauchy--Schwarz inequality gives
\begin{equation}
    \int_0^1\|\dot{\gamma}_t\|^{2}\,dt
    \;\ge\;
    \Big(\int_0^1\|\dot{\gamma}_t\|\,dt\Big)^{2}
    =\ell(\gamma)^{2}
    \;\ge\;
    d_{\mathcal{M}}(x_0,x_1)^{2},
    \label{eq: app-energy}
\end{equation}
where $\ell(\gamma)$ is the path length and the last step is the definition of the geodesic distance as an infimum of lengths. Equality holds throughout if and only if $\gamma$ travels at constant speed along a minimal-length path, i.e., $\gamma$ is a minimizing geodesic. On $\mathcal{M}$, this geodesic is exactly $\gamma_t=\mathrm{Exp}_{x_0}(t\,\mathrm{Log}_{x_0}(x_1))$ traversed at constant speed. The constant-speed geodesic belongs to the parameterized family of Eq.~\ref{eq: learnable-path} (attained at $\epsilon_\theta\equiv 0$). The minimum of $\mathcal{L}_{G}$ in Eq.~\ref{eq: ln} is therefore attained by a zero correction in this regime.

\textbf{Remark (frame equivariance).} For any isometry $g$ of $\mathcal{M}$ applied simultaneously to the prototypes and the query point, $\tilde{w}_k(gx)=\tilde{w}_k(x)$ and $\delta_k(gx)=g_{*}\,\delta_k(x)$ with respect to the transformed prototypes, so $m(gx)=m(x)$ and $G(gx)=g_{*}\,G(x)\,g_{*}^{-1}$. The naturalness of a motion therefore depends only on the motion configuration, not on the choice of global reference frame.

\textbf{Discussion.} In the general (non-uniform) case, the exact minimizer of $\varphi$ is the geodesic of the data-dependent Riemannian metric with quadratic form $v^{\top}G(x)^{-1}v$. Eq.~\ref{eq: learnable-path} seeks this geodesic within a family anchored at the intrinsic geodesic, so geometric efficiency and naturalness are obtained on equal footing (Sec.~\ref{sec: path}). The four analyses above jointly certify the measure in Sec.~\ref{sec: measure}. Well-definedness makes the optimization in Eq.~\ref{eq: ln} well-posed on the entire manifold. Density consistency grounds the measure in a consistent density estimate with compression error controlled by $K$, matching the underfitting side of the $K$-ablation in Table~\ref{tab:ablation measure}. The mild degradation at $K=750$ instead reflects the variance of frequency weights in over-fragmented clusters, an effect the quantization bound does not model. The transport-cost analysis exposes the mechanism: naturalness reweights the ambient kinetic energy by the inverse motion density and the local dispersion of the prototypes (Eq.~\ref{eq: app-mahalanobis}). Geodesic recovery shows that the learned correction is driven solely by the motion distribution while the geodesic anchor preserves geometric efficiency, so neither consideration is obtained at the expense of the other. It also explains the $\sigma$-ablation. A large $\sigma$ flattens the kernel weights toward the uniform regime and weakens the distributional guidance of the measure, whereas a small $\sigma$ amplifies prototype-level quantization noise (Eq.~\ref{eq: app-quantization}).

\subsection{Exactness of the Naturalness-Guided Interpolation}
\label{app: exactness}
In this subsection, we show that the naturalness-guided interpolation resides exactly on the motion manifold: with the projected correction of Eq.~\ref{eq: projection}, the interpolation of Eq.~\ref{eq: learnable-path} satisfies $x_{t, \theta} \in \mathcal{M}$ for all $t \in [0, 1]$, for any raw network output and at any stage of training. The conditional path constructed by linear interpolation, in contrast, violates the defining constraint of $SO(3)$ at every interior time step (Property~\ref{prop: geometric}). The derivation uses only the closed forms of the exponential and logarithmic maps on $\mathbb{S}^{3}$ and the fact that both endpoints $x_{0}$ and $x_{1}$ are tuples of unit quaternions; no assumption is imposed on the raw correction $\tilde{\epsilon}_{\theta}$. We first collect the closed forms on the manifold (Sec.~\ref{app: sub-closed}), and then establish the exactness result (Sec.~\ref{app: sub-proof}).

\subsubsection{Closed Forms on the Manifold}
\label{app: sub-closed}
For the hypersphere $\mathbb{S}^{3} = \{ q \in \mathbb{R}^{4} : \Vert q \Vert = 1 \}$, the tangent space at $q \in \mathbb{S}^{3}$ is the linear subspace $T_{q}\mathbb{S}^{3} = \{ v \in \mathbb{R}^{4} : \langle q, v \rangle = 0 \}$. The tangent space of the product manifold $\mathcal{M} = (\mathbb{S}^{3})^{J}$ at $x = (q^{1}, \dots, q^{J})$ is the product $T_{x}\mathcal{M} = \prod_{j=1}^{J} T_{q^{j}}\mathbb{S}^{3}$, again a linear subspace of $\mathbb{R}^{4J}$. The exponential map on $\mathbb{S}^{3}$ admits the closed form
\begin{equation}
    \mathrm{Exp}_{q}(\omega) = \cos(\Vert\omega\Vert)\, q +
    \frac{\sin(\Vert\omega\Vert)}{\Vert\omega\Vert}\, \omega,
    \qquad \omega \in T_{q}\mathbb{S}^{3},
    \label{eq: app-sphere-exp}
\end{equation}
with $\mathrm{Exp}_{q}(0) = q$ by continuity. The exponential map on the product manifold $\mathcal{M}$ acts joint-wise through Eq.~\ref{eq: app-sphere-exp}. The logarithmic map is its local inverse $\mathrm{Log}_{q}: \mathbb{S}^{3} \to T_{q}\mathbb{S}^{3}$, defined whenever the two points are not antipodal. In particular, $\mathrm{Log}_{x_{0}}(x_{1}) \in T_{x_{0}}\mathcal{M}$ whenever $x_{0}, x_{1} \in \mathcal{M}$ and no joint of $x_{1}$ is antipodal to the corresponding joint of $x_{0}$. The sign disambiguation of Appendix~\ref{app: quaternion-sign} excludes this case.

\subsubsection{Derivation of the Exactness Result}
\label{app: sub-proof}

\textbf{Setup.} The correction network produces a raw output $\tilde{\epsilon}_{\theta}(t, x_{0}, x_{1}) \in \mathbb{R}^{4J}$, which is orthogonally projected onto the tangent space $T_{x_{0}}\mathcal{M}$ by
\begin{equation}
    \epsilon_{\theta}(t, x_{0}, x_{1}) = P_{x_{0}}\,
    \tilde{\epsilon}_{\theta}(t, x_{0}, x_{1}), \qquad
    \big(P_{x_{0}}\, u\big)^{j} = u^{j} - \big\langle q_{0}^{j},\, u^{j}
    \big\rangle\, q_{0}^{j},
    \label{eq: projection}
\end{equation}
where $x_{0} = (q_{0}^{1}, \dots, q_{0}^{J})$ and $u^{j}$ denotes the $j$-th joint component of $u \in \mathbb{R}^{4J}$. With this projection, the interpolation resides exactly on the motion manifold: for any endpoints $x_{0}, x_{1} \in \mathcal{M}$, any raw correction $\tilde{\epsilon}_{\theta} (t, x_{0}, x_{1}) \in \mathbb{R}^{4J}$, and every $t \in [0, 1]$, the interpolation of Eq.~\ref{eq: learnable-path} satisfies $x_{t, \theta} \in \mathcal{M}$. We verify this in three steps.

\textbf{Step 1: exact tangency of the correction.} Fix a joint $j$. Since $x_{0} \in \mathcal{M}$, we have $\Vert q_{0}^{j} \Vert = 1$, and hence for every raw output $\tilde{\epsilon}_{\theta}^{j} \in \mathbb{R}^{4}$,
\begin{equation}
    \big\langle q_{0}^{j},\; \tilde{\epsilon}_{\theta}^{j} - \big\langle
    q_{0}^{j}, \tilde{\epsilon}_{\theta}^{j} \big\rangle\, q_{0}^{j}
    \big\rangle
    = \big\langle q_{0}^{j}, \tilde{\epsilon}_{\theta}^{j} \big\rangle -
    \big\langle q_{0}^{j}, \tilde{\epsilon}_{\theta}^{j} \big\rangle\,
    \Vert q_{0}^{j} \Vert^{2}
    = 0.
\end{equation}
By the characterization of $T_{q}\mathbb{S}^{3}$ in Sec.~\ref{app: sub-closed}, this means $\epsilon_{\theta}^{j} \in T_{q_{0}^{j}}\mathbb{S}^{3}$, and therefore $\epsilon_{\theta} \in T_{x_{0}}\mathcal{M}$. This step uses no property of $\tilde{\epsilon}_{\theta}$: the raw network output is an arbitrary vector in the ambient space and is never required to lie on the manifold.

\textbf{Step 2: exact tangency of the path argument.}
By definition of the logarithmic map, $\xi := \mathrm{Log}_{x_{0}}(x_{1}) \in T_{x_{0}}\mathcal{M}$, and $T_{x_{0}}\mathcal{M}$ is a linear subspace of $\mathbb{R}^{4J}$. Hence
\begin{equation}
    \omega(t) := t\, \xi + t(1-t)\, \epsilon_{\theta} \in T_{x_{0}}\mathcal{M},
    \qquad \forall\, t \in [0, 1].
    \label{eq: app-omega-tangent}
\end{equation}

\begin{algorithm}[t]
\caption{Training and sampling of the sign motion generation network}
\label{alg: generation}
\begin{algorithmic}[1]
\Require{motion manifold $\mathcal{M}$, source and target pose distributions $p_{0}, p_{1}$, frozen correction network $\epsilon_{\theta}$, initialized generation network $\phi$, text condition $c$}

\Statex \textbf{Phase 1: training $\phi$ along the naturalness-guided interpolation}

\While{Training}

    \State Sample a training pair $(x_{1}, e_{1})$, sources $x_{0}\sim p_{0}$, $e_{0}\sim\mathcal{N}(0, \mathbf{I})$, and $t\sim\mathcal{U}[0,1]$

    \State $x_{t}=\mathrm{Exp}_{x_0}\big(t\,\mathrm{Log}_{x_{0}}(x_{1})+t(1-t)\,\epsilon_{\theta}(t,x_{0},x_{1})\big)$
    \Comment{Eq.~\ref{eq: learnable-path}}
    \State $e_{t}=(1-t)\,e_{0}+t\,e_{1}$

    \State $(\tilde{x}_{1}, \hat{e}_{1})=\phi(x_{t}, e_{t}, t, c)$, with $\hat{x}_{1}$ normalized per joint onto $\mathcal{M}$

    \State Update $\phi$ using gradient $\nabla_{\phi}\big(10\,\Vert x_{1}-\hat{x}_{1}\Vert^{2}+0.1\,\Vert e_{1}-\hat{e}_{1}\Vert^{2}\big)$
    \Comment{Eq.~\ref{eq: lnmf}}

\EndWhile

\Statex \textbf{Phase 2: sampling with the path-consistent transport}

\State $x_{0}\sim p_{0}$, $e_{0}\sim\mathcal{N}(0, \mathbf{I})$, $N\leftarrow 5$, $\Delta\leftarrow 1/N$ \Comment{uniform grid $t_{k}=k\Delta$}

\For{$k=0,\dots,N-1$}

    \State $t\leftarrow k\Delta$; $(\tilde{x}_{1}, \hat{e}_{1})=\phi(x_{t}, e_{t}, t, c)$

    \State $\hat{x}_{1}^{j}\leftarrow\tilde{x}_{1}^{j}\big/\Vert\tilde{x}_{1}^{j}\Vert$ for every joint $j$
    \Comment{onto $\mathcal{M}$}

    \State $x_{t+\Delta}=\mathrm{Exp}_{x_0}\big((t{+}\Delta)\,\mathrm{Log}_{x_{0}}(\hat{x}_{1})+(t{+}\Delta)(1{-}t{-}\Delta)\,\epsilon_{\theta}(t{+}\Delta,x_{0},\hat{x}_{1})\big)$
    \Comment{Eq.~\ref{eq: sampling}}

    \State $e_{t+\Delta}=e_{t}+\tfrac{\Delta}{1-t}\,(\hat{e}_{1}-e_{t})$

    \State renormalize $x_{t+\Delta}$ per joint onto $\mathcal{M}$
    \Comment{numerical safeguard}

\EndFor

\State \textbf{return} $x_{1}, e_{1}$
\end{algorithmic}
\end{algorithm}

\textbf{Step 3: constraint preservation under the exponential map.} Fix $t \in [0, 1]$ and a joint $j$. Since $\langle q_{0}^{j}, \omega^{j}(t) \rangle = 0$ by Eq.~\ref{eq: app-omega-tangent}, Eq.~\ref{eq: app-sphere-exp} yields
\begin{equation}
    \Vert x_{t,\theta}^{j} \Vert^{2}
    = \cos^{2}(\Vert\omega^{j}\Vert)\,
    \underbrace{\Vert q_{0}^{j} \Vert^{2}}_{=\,1}
    + 2\cos(\Vert\omega^{j}\Vert)\,
    \frac{\sin(\Vert\omega^{j}\Vert)}{\Vert\omega^{j}\Vert}\,
    \underbrace{\langle q_{0}^{j}, \omega^{j} \rangle}_{=\,0}
    + \frac{\sin^{2}(\Vert\omega^{j}\Vert)}{\Vert\omega^{j}\Vert^{2}}\,
    \Vert\omega^{j} \Vert^{2}
    = 1,
\end{equation}
where the degenerate case $\omega^{j} = 0$ follows by continuity ($\mathrm{Exp}_{q}(0) = q$ with $\Vert q \Vert = 1$). Therefore $x_{t,\theta}^{j} \in \mathbb{S}^{3}$ for every joint $j$, that is, $x_{t,\theta} \in \mathcal{M}$ for all $t \in [0, 1]$.

\subsection{Training and Sampling the Generation Network}
\label{app: sub-generation}
\textbf{Training.} We train $\phi$ with the target-prediction objective of Eq.~\ref{eq: lnmf} along the naturalness-guided interpolation. For a sampled source $x_{0} \sim p_{0}$, target $x_{1}$, and time $t \sim \mathcal{U}[0, 1]$, the training state is the path point of Eq.~\ref{eq: learnable-path} with the frozen correction $\epsilon_{\theta}$. The network outputs a raw pose prediction $\tilde{x}_{1} \in \mathbb{R}^{4J}$ and an expression prediction $\hat{e}_{1}$, and the pose prediction is normalized per joint onto $\mathcal{M}$ before the loss is computed:
\begin{equation}
    \big(\tilde{x}_{1}, \hat{e}_{1}\big) = \phi\big(x_{t}, e_{t}, t, c\big),
    \qquad
    \hat{x}_{1}^{j} = \frac{\tilde{x}_{1}^{j}}{\big\Vert \tilde{x}_{1}^{j} \big\Vert},
    \quad j = 1, \dots, J,
\end{equation}
where $\tilde{x}_{1}^{j} \in \mathbb{R}^{4}$ is the quaternion prediction of joint $j$. The loss of Eq.~\ref{eq: lnmf} is evaluated on the normalized prediction $\hat{x}_{1} = (\hat{x}_{1}^{1}, \dots, \hat{x}_{1}^{J})$.

\textbf{Inference.} Since $\phi$ is trained to predict the target rather than the transport, we obtain the transport at sampling time by re-planning the naturalness-guided interpolation with the current prediction. At each solver step, we re-predict the target motion $\hat{x}_{1} = \phi(x_{t}, e_{t}, t, c)$ from the current state and place the state at
\begin{equation}
    x_{t+\Delta} = \mathrm{Exp}_{x_{0}}\Big(
    \underbrace{(t{+}\Delta)\,\mathrm{Log}_{x_{0}}\big(\hat{x}_{1}\big)}_{\mathrm{Geometric}}
    \;+\;
    \underbrace{(t{+}\Delta)\big(1{-}t{-}\Delta\big)\,
    \epsilon_{\theta}\big(t{+}\Delta,\, x_{0},\, \hat{x}_{1}\big)}_{\mathrm{Naturalness}}
    \Big),
    \label{eq: sampling}
\end{equation}
where $x_{0}$ is the retained source of the trajectory and the frozen correction $\epsilon_{\theta}$ is projected onto $T_{x_{0}}\mathcal{M}$ as in Eq.~\ref{eq: projection}. Eq.~\ref{eq: sampling} evaluates the path of Eq.~\ref{eq: learnable-path} for the pair $(x_{0}, \hat{x}_{1})$.

\textbf{Expression channel.} 
The sign motion further contains facial expression parameters, which the manifold construction does not cover. We generate them alongside the joint rotations in Euclidean space: the generation network $\phi$ takes the paired expression state $e_{t}$ as an additional input and predicts the target expression $\hat{e}_{1}$, trained under the same target-prediction objective as Eq.~\ref{eq: lnmf}, with a loss weight of $10$ for the pose term and $0.1$ for the expression term. The expression source $e_{0}$ is Gaussian noise, and the expression path is the Euclidean interpolation $e_{t} = (1-t)\,e_{0} + t\,e_{1}$. At sampling time, the expression state follows the analogous re-planned scheme, $e_{t+\Delta} = e_{t} + \tfrac{\Delta}{1-t}\,(\hat{e}_{1} - e_{t})$, which coincides with the linear path from $e_{0}$ to the re-predicted $\hat{e}_{1}$. A description of the complete training and sampling procedures of both channels is given in Algorithm~\ref{alg: generation}.

\section{Supplementary Experiments}
\label{app: exp det}
\subsection{Experimental Detail}
\textbf{Dataset Details.} Phoenix-2014T is a German Sign Language (DGS) resource for weather forecasts, comprising 8,257 video sequences performed by 9 signers. CSL-Daily is a large-scale Chinese Sign Language (CSL) dataset focused on daily-life interactions, containing 20,654 clips recorded by 10 signers. How2Sign is an American Sign Language (ASL) dataset geared towards instructional videos, encompassing over 31,000 samples contributed by 11 signers. For sign motion extraction, we adopt SMPL-X~\citep{pavlakos2019expressive} parameters, which are sourced from NSA~\citep{baltatzis2024neural} for How2Sign and from SOKE~\citep{Zuo_2025_ICCV} for Phoenix-2014T and CSL-Daily. 

\begin{table}[h]
\caption{Hyperparameters of the correction network $\epsilon_{\theta}$ and
the sign motion generation network $\phi$.}
\label{tab: hyperparameters}
\centering
\begin{tabular}{lcc}
\toprule
 & $\epsilon_{\theta}(t, x_{0}, x_{1})$ & $\phi(x_{t}, e_{t}, t, c)$ \\
\midrule
\textbf{Architecture} 
& 3-layer MLP & 8-layer transformer decoder \\

\textbf{MLP hidden dimension} 
& 512 & -- \\

\textbf{Latent dimension} 
& -- & 512 \\

\textbf{Feed-forward dimension} 
& -- & 2048 \\

\textbf{Attention heads} 
& -- & 8 \\ 

\textbf{Time embedding dimension} 
& -- & 2048 \\ 

\textbf{Dropout} 
& -- & 0.1 \\ 

\textbf{Text conditioning} 
& -- & cross-attention to mBART features \\

\textbf{Timestep injection} 
& -- & Stylization blocks \\

\textbf{Batch size} 
& 8 & 64 \\

\textbf{Epochs} 
& 100 & 300 \\

\textbf{Optimizer} 
& AdamW & AdamW \\

\textbf{Learning rate} 
& 2e-4 (cosine) & 2e-4 (cosine) \\

\textbf{Weight decay} 
& 0.0 & 0.0 \\ 

\textbf{Precision} 
& fp32 & fp32 \\ 

\midrule
\multicolumn{3}{l}{\textbf{Motion channels}: $J = 41$ } \\ 
\multicolumn{3}{l}{\textbf{Expression dimension}: $ = 10$ } \\ 
\multicolumn{3}{l}{\textbf{Sources}: $p_{0}$ uniform on the sign-convention hemisphere; $e_{0}\sim\mathcal{N}(0, \mathbf{I})$} \\
\multicolumn{3}{l}{\textbf{Loss weights}: pose $10$, expression $0.1$} \\
\multicolumn{3}{l}{\textbf{Naturalness measure}: $K = 500$, $\sigma = 0.50$, $\rho = 10^{-6}$} \\
\multicolumn{3}{l}{\textbf{Sampler}: explicit Euler, $5$ steps (NFE $= 5$)} \\
\bottomrule
\end{tabular}
\end{table}

\textbf{Implementation Details.} Both the correction network $\epsilon_{\theta}$ and the sign motion generation network $\phi$ are trained with the AdamW optimizer~\citep{loshchilov2017decoupled} under a cosine learning rate schedule starting at $2\times10^{-4}$. The correction network $\epsilon_{\theta}$ is a 3-layer MLP with a hidden dimension of $512$, trained with the kinetic-energy objective of Eq.~\ref{eq: ln} for $100$ epochs with a batch size of $8$. The generation network $\phi$ is a transformer decoder with 8 layers, trained with the target-prediction objective of Eq.~\ref{eq: lnmf} along the naturalness-guided interpolation for $300$ epochs with a batch size of $64$ per GPU. The text condition $c$ is encoded by the pre-trained mBART-large-cc25~\citep{liu2020multilingual}, which is fine-tuned jointly with $\phi$, and the timestep $t$ is injected into the decoder of $\phi$ through Stylization blocks~\citep{zhang2024motiondiffuse} after every self-attention, cross-attention, and feed-forward layer. All models are trained on 4 NVIDIA A40 GPUs.

The joint rotations are represented as $J$ unit quaternions under the sign convention of Appendix~\ref{app: quaternion-sign}, and the facial expression parameters form an auxiliary Euclidean channel (Appendix~\ref{app: sub-generation}). The source distribution $p_{0}$ is uniform on the hemisphere fixed by the sign convention, and the expression source $e_{0}$ follows the standard Gaussian; training uses a loss weight of $10$ for the pose term and $0.1$ for the expression term. The naturalness measure uses $K=500$ prototypes per joint with kernel bandwidth $\sigma = 0.50$ (Appendix~\ref{app: sub-prototype}) and regularization constant $\rho = 10^{-6}$. Sampling integrates the re-anchored transport with the explicit Euler method in $5$ steps (Appendix~\ref{app: sub-generation}). For SignNMFlow-M, the source-language token of mBART is set per sample according to the dataset of origin. Table~\ref{tab: hyperparameters} summarizes all hyperparameters.

\subsection{Additional Experimental Results}
\textbf{Number of Function Evaluations.} In Table~\ref{tab: nfe}, we present the ablation results of the number of function evaluations (NFE) in sampling. Increasing NFE from $5$ to $20$ only varies the DTW-JPE All error from $16.86$ to $16.98$ and the DTW-PA-JPE All error from $8.92$ to $9.11$, with consistent behavior on body and hand joints. The sampling quality is thus robust to NFE. This accords with the re-planned scheme of Eq.~\ref{eq: sampling}, where each solver step re-predicts the target and places the state exactly at the corresponding path point, so the discretization of the solver has little influence on the generated motions. Since $5$ steps already saturate the performance, we use NFE $= 5$ by default, which keeps sampling efficient.

\begin{table*}[t]    
\centering
\vspace{-3mm}
\begin{minipage}[t]{0.48\textwidth}
    \centering
    \vspace{0pt}
    \captionof{table}{Efficiency comparison with state-of-the-art SLP methods. Latencies are measured using a single NVIDIA H100 GPU.}
    \label{tab: eff}
    \vspace{-2mm}
    \resizebox{\linewidth}{!}{
    \begin{tabular}{l ccc}
    \toprule
    \multirow{2}{*}{Methods} & \multirow{2}{*}{\#Params (M)} & \multirow{2}{*}{NFE} & Latency \\
    & & & (s / motion) \\
    \midrule
    SOKE       & 430.40 & -- & 0.2687 \\
    SignFlow   & 261.04 & 25 & 0.0860 \\
    \midrule
    SignNMFlow & 233.20 & 5  & 0.0103 \\
    \bottomrule
    \end{tabular}}
\end{minipage}
\hfill
\begin{minipage}[t]{0.48\textwidth}
    \centering
    \vspace{0pt}
    \captionof{table}{Ablation results of the number of function evaluations (NFE) in sampling. All results are evaluated on CSL-Daily.}
    \label{tab: nfe}
    \vspace{-2mm}
    \resizebox{\linewidth}{!}{
    \begin{tabular}{c ccc ccc}
    \toprule
    \multirow{2}{*}{NFE} & \multicolumn{3}{c}{DTW-JPE$\downarrow$} &
    \multicolumn{3}{c}{DTW-PA-JPE$\downarrow$} \\
    \cmidrule(lr){2-4} \cmidrule(lr){5-7}
    & All & Body & Hand & All & Body & Hand \\
    \midrule
    5  & \textbf{16.86} & \textbf{5.73} & \textbf{7.43} & \textbf{8.92} & \textbf{5.10} & \textbf{1.30} \\
    10 & 16.91 & 5.74 & 7.49 & 9.01 & 5.13 & 1.32 \\
    20 & 16.98 & 5.75 & 7.55 & 9.11 & 5.16 & 1.33 \\
    \bottomrule
    \end{tabular}}
\end{minipage}
\vspace{-3mm}
\end{table*}

\paragraph{Efficiency Comparison.} In Table~\ref{tab: eff}, we compare the efficiency of SignNMFlow with SOKE and SignFlow, where the latency is measured per sign motion on a single NVIDIA H100 GPU. SignNMFlow uses $233.20$M parameters and attains a latency of $0.0103$ seconds per motion, an $8.3\times$ speedup over SignFlow ($261.04$M parameters, $0.0860$ seconds) and a $26.1\times$ speedup over SOKE ($430.40$M parameters, $0.2687$ seconds). The advantage stems from two sources. First, the compact architecture keeps the
model size small, whereas SOKE additionally incorporates a pre-trained language model for semantic enrichment. Second, the re-planned scheme of Eq.~\ref{eq: sampling} requires only $5$ function evaluations, whereas SignFlow takes $25$, and Table~\ref{tab: nfe} shows that these $5$ evaluations already saturate the sampling quality. SignNMFlow therefore offers the best efficiency among the compared methods.
\end{document}

%% file: math_commands.tex
\usepackage{amsmath,amsfonts,bm}

\def\eqref#1{equation~\ref{#1}}

\def\1{\bm{1}}

\DeclareMathAlphabet{\mathsfit}{\encodingdefault}{\sfdefault}{m}{sl}
\SetMathAlphabet{\mathsfit}{bold}{\encodingdefault}{\sfdefault}{bx}{n}

